\documentclass[preprint,12pt,authoryear]{elsarticle}

\usepackage[utf8]{inputenc} 
\usepackage[T1]{fontenc}    
\usepackage{hyperref}       
\usepackage{url}            
\usepackage{booktabs}       
\usepackage{amsfonts}       
\usepackage{nicefrac}       
\usepackage{microtype}      
\usepackage{xcolor}         
\usepackage{graphicx}
\usepackage{algorithm}
\usepackage{algpseudocode}
\usepackage[normalem]{ulem}
\usepackage{amsmath}
\usepackage{multirow}
\usepackage{soul}
\usepackage{subcaption}  
\usepackage{placeins}
\usepackage{wrapfig}
\usepackage{float}

\journal{Medical Image Analysis}

\begin{document}

\begin{frontmatter}


\title{NucEval: Revisiting and Extending Evaluation Metrics for Nuclear Instance Segmentation}

\author[1,2]{Amirreza Mahbod\corref{cor1}}
\ead{amirreza.mahbod@dp-uni.ac.at}
\cortext[cor1]{Corresponding author}

\author[1]{Ramona Woitek}

\author[2,3]{Jeanne Shen}

\affiliation[1]{organization={Research Center for Medical Image Analysis and Artificial Intelligence, Department of Medicine, Faculty of Medicine and Dentistry, Danube Private University},
            city={Krems an der Donau},
            postcode={3500},
            country={Austria}}

\affiliation[2]{organization={Department of Pathology, Stanford University School of Medicine},
            city={Stanford},
            postcode={94305},
            state={CA},
            country={USA}}

\affiliation[3]{organization={Center for Artificial Intelligence in Medicine \& Imaging, Stanford University},
            city={Stanford},
            postcode={94305},
            state={CA},
            country={USA}}

\begin{abstract}

In computational pathology, nuclear instance segmentation is a fundamental task with many downstream clinical applications. With the advent of deep learning, many approaches, including convolutional neural networks (CNNs) and vision transformers (ViTs), have been proposed for this task, along with both machine learning-based and non-machine learning-based pre- and post-processing techniques to further boost performance. However, one fundamental aspect that has received less attention is the evaluation pipeline.
In this study, we identify four key issues associated with nuclear instance segmentation evaluation and propose corresponding solutions. Our proposed modifications, namely handling vague regions, score normalization, overlapping instances, and border uncertainty, are integrated into a unified framework called NucEval, which extends commonly used metrics for evaluating nuclear instance segmentation.
We evaluate this framework on the NuInsSeg dataset, which provides unique characteristics that make it particularly suitable for this study, as well as two additional external datasets, using three CNN- and ViT-based nuclear instance segmentation models. Beyond its effect on the metrics, we show that NucEval increases inter- and intra-observer agreement between expert annotations and can change the ranking of closely competing methods on an external benchmark, indicating that the choice of evaluation protocol can affect model selection.
The code, along with complete guidelines and illustrative examples, is publicly available at: \url{https://github.com/masih4/nuc_eval}.

\end{abstract}



\begin{keyword}

Nuclear Instance Segmentation \sep
Evaluation Metrics \sep
Deep Learning \sep
Computational Pathology \sep
Medical Imaging 
\end{keyword}

\end{frontmatter}

\section{Introduction}
\label{sec:intro}
Histologic image analysis is considered the gold standard for diagnosis and prognostication of many diseases, including different types of cancer~\citep{jmp7010002}. Among the multiple tasks comprising  automated histologic image analysis, nuclear analysis is regarded as one of the most fundamental, as it provides essential information for many downstream clinical applications~\citep{McGenity2024, https://doi.org/10.1049/ipr2.70129, monuseg}. Since manual segmentation is infeasible for whole-slide images (WSIs), which may contain up to one million nuclei~\citep{hou2020dataset}, computerized approaches are commonly employed. The computational framework for nuclear instance segmentation generally consists of three main components, including datasets for training and testing, segmentation models, and evaluation components~\citep{torbati2026nucfuserank, gamper2020pannuke, graham2021lizard, Nasir2023, NUNES2025103360}.

Most research has focused on improving segmentation models, including CNN-based architectures such as HoVer-Net~\citep{graham2019hover} and HoVer-NeXt~\citep{pmlr-v250-baumann24a} and ViT-based models such as CellViT~\citep{HORST2024103143}, as well as on developing pre-processing techniques such as stain normalization~\citep{10.1007/978-3-031-16434-7_14, mahbod2024improving} and post-processing strategies such as watershed-based instance separation~\citep{ mahbod2024improving,lv2025kongnet}. However, the evaluation pipeline has received comparatively less attention. The lack of reliable evaluation strategies may result in suboptimal performance estimation and hinder fair comparison between methods~\citep{Muuller2022, Reinke2024}, potentially leading to the selection of inferior models for downstream clinical applications.

A variety of metrics have been used in the literature across different studies and challenges related to nuclear instance segmentation (and classification). Some metrics, such as the Dice score~\citep{ed278621-dc3e-343f-ae66-540d8990b60d} or the Jaccard index~\citep{jaccard1901distribution}, were originally designed for semantic segmentation~\citep{Muuller2022}. Although these metrics are also commonly reported for nuclear instance segmentation, they are typically used as secondary metrics, as they do not adequately capture instance-level performance, particularly the ability of a model to separate overlapping nuclei~\citep{monuseg}. In some studies and challenges, such as the PUMA challenge~\citep{schuiveling2025novel, Schuiveling2026.03.09.26347935} or nuclear instance segmentation benchmarking for immunofluorescence images~\citep{Sankaranarayanan2025},the task is based on nuclear instance segmentation, but detection-based scores such as instance-level F1-scores are also reported. Despite the variety of metrics employed, the panoptic quality (PQ)~\citep{Kirillov_2019_CVPR} and aggregated Jaccard index (AJI)~\citep{monuseg} have emerged as the most widely adopted primary metrics for nuclear instance segmentation, as they jointly evaluate both the detection and segmentation quality of individual instances. The formal definitions and mathematical formulations of PQ, which is decomposed into detection quality (DQ) and segmentation quality (SQ), as well as AJI, are provided in their respective original studies~\citep{Kirillov_2019_CVPR, monuseg}. However, several issues with current evaluation practices have been identified in the literature. For example, an evaluation issue was identified in the implementation of the MoNuSAC challenge metric~\citep{9745980}. Although this bug did not affect the overall challenge ranking~\citep{9745890}, it highlights the importance of carefully designing and validating the evaluation process. Other studies have discussed common limitations of segmentation metrics, including those widely used for instance segmentation in medical imaging~\citep{graham2019hover, reinke2021common, Foucart2023, Maier-Hein2024}.

Beyond metric design, evaluation challenges have also been discussed in the broader context of manual annotation in medical and histologic image analysis. Several studies have reported that manual segmentation masks, which are typically considered ground-truth annotations, can be error-prone and subject to low inter- and intra-observer
agreement~\citep{FOUCART2023102155, 10.1007/978-3-030-00928-1_77}. For nuclear instance segmentation, for example, the inter- and intra-observer variability for the CryoNuSeg dataset, measured using Dice, AJI, and PQ, was reported as 78.9\%, 54.1\%, and 50.9\%, and 83.8\%, 62.4\%, and 57.0\%, respectively~\citep{mahbod2021cryonuseg_org}.

In addition to the choice of metric, the manner in which the metric is applied can also introduce bias into the evaluation pipeline. For evaluation, a set of samples in the so-called ``test set'' is typically used. In some cases, the metric is computed per image in the test set, and the final score is obtained by averaging across samples, as done in challenges such as MoNuSAC, MoNuSeg, or the nuclear segmentation track of the PUMA challenge~\citep{monuseg, schuiveling2025novel, 9446924}. In other cases, all test images are merged into a single large test sample and the score is computed globally over the merged test sample, as in the CoNIC challenge and the tissue segmentation track of the PUMA challenge~\citep{schuiveling2025novel, graham2023conic}. Issues related to robustness and aggregation methods used for ranking algorithms have also been reported in other medical image analysis challenges, further highlighting the importance of this problem~\citep{Maier-Hein2018}. For example, as reported by \citet{Maier-Hein2018}, weaknesses in evaluation and ranking strategies can be exploited to drastically alter the ranking of participating teams in medical image segmentation tasks~\citep{10.1007/978-3-030-00937-3_45}. Therefore, precautions such as reporting rankings with confidence intervals, conducting on-site evaluations, or publicly releasing the evaluation software should be considered to ensure fairness and transparency~\citep{10.1007/978-3-030-00937-3_45, foucart2025ranking}.


In this study, we consider issues in the evaluation pipeline that may arise from different sources, such as inter- and intra-observer variability and bias toward specific samples in the test set. We identify four main limitations in current evaluation practices, namely issues associated with vague areas, unnormalized scores, overlapping areas, and uncertainty in the annotation borders, and propose solutions to address them within the evaluation pipeline. Our modifications can be integrated into commonly used evaluation metrics. Specifically, we introduce NucEval, a ready-to-use Python-based framework that incorporates these modifications into widely used metrics for assessing semantic and instance-based nuclear segmentation performance, including PQ, AJI, Dice, DQ, and SQ. We demonstrate the effect of the proposed modifications using well-known, publicly available benchmark datasets, namely NuInsSeg~\citep{mahbod2024nuinsseg} (as the main evaluation dataset), CryoNuSeg~\citep{mahbod2021cryonuseg_org}, and PCNS~\citep{hou2020dataset}, together with three state-of-the-art models, namely HoVer-Net~\citep{graham2019hover}, HoVer-NeXt~\citep{pmlr-v250-baumann24a}, and CellViT~\citep{HORST2024103143}. In addition, we assess whether the modifications bring the evaluation closer to expert consensus by measuring inter- and intra-observer agreement on CryoNuSeg, and whether they affect model selection by re-evaluating top-performing methods from the MoNuSAC challenge~\citep{9446924}. We make our code for NucEval and the modified metrics publicly available to support further research in this area.

\section{Methods}
\label{sec:me}
\begin{figure}[ht]
	\centering
	\includegraphics[width=0.95\linewidth]{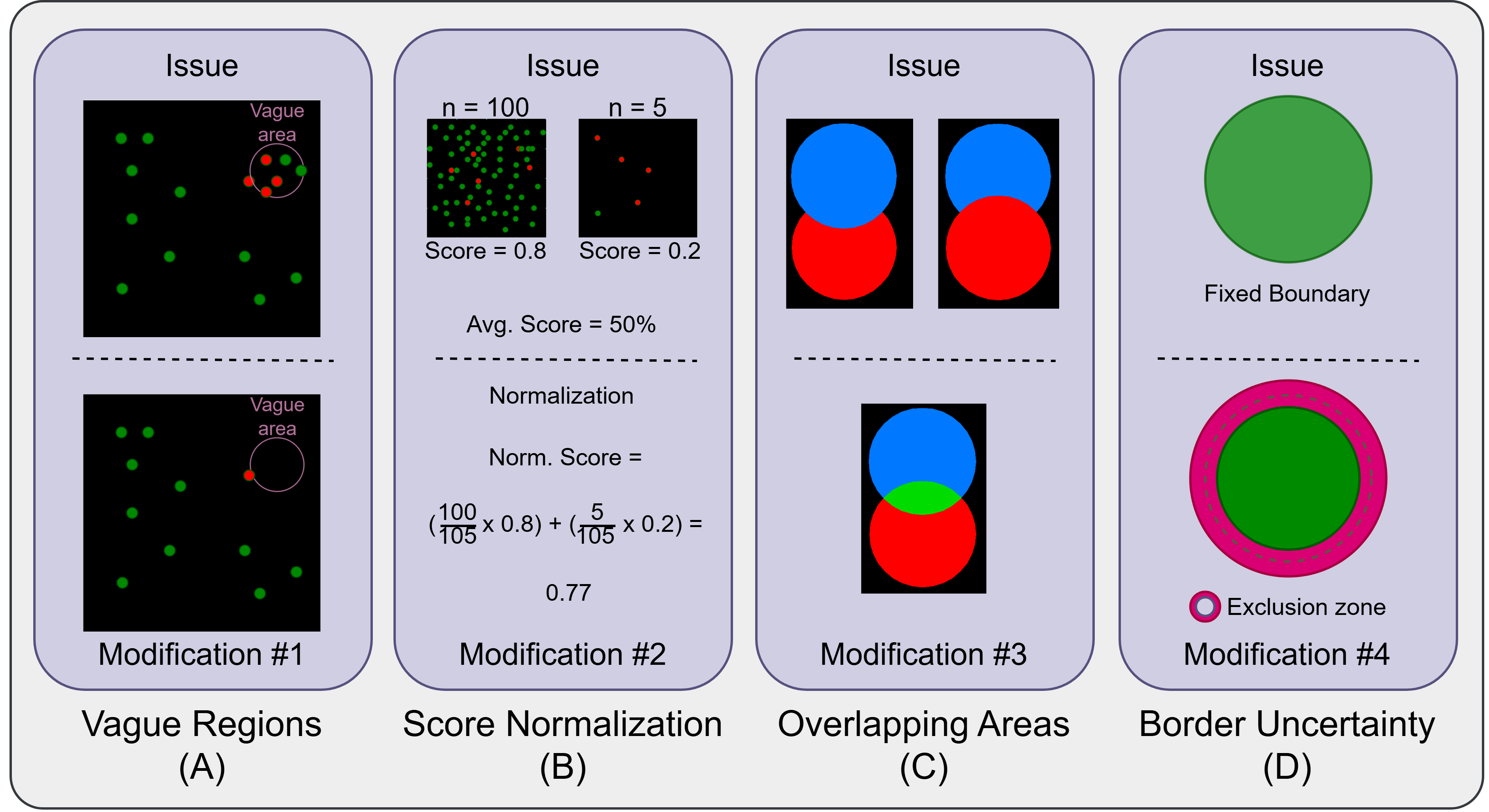}
	\caption{Overview of the four proposed modifications (lower part of each box) for the identified issues (upper part of each box), including (A) incorporation of vague areas into the evaluation by excluding them from the evaluation pipeline, (B) score normalization based on nucleus count, where images with more nuclei are given higher weight when computing the overall score, (C) handling of overlapping regions by assigning the overlapped area to all instances that share it, instead of assigning it to only one of the instances, and (D) border delineation uncertainty handling by creating a zone around the annotation border and excluding this zone from the evaluation. The images and scores are for illustrative purposes only and do not represent actual values; they are intended solely to convey the core idea of each component. }
	\label{fig:solution}
\end{figure}

The main components of the identified issues and their corresponding proposed solutions are illustrated schematically in Figure~\ref{fig:solution}. A detailed description of each component is provided in Sections~\ref{sec:vague areas}, \ref{sec:score_normalization}, \ref{sec:ovelapped} and \ref{sec:border_uncertain}.

\subsection{Dataset}
\begin{figure}
	\centering
	\begin{tabular}{cccc}
		\small Tissue & \small Labeled mask & \small Labeled mask & \small Vague mask \\
		& \small (preserved) & \small (merged) &  \\
		\includegraphics[width=0.15\textwidth]{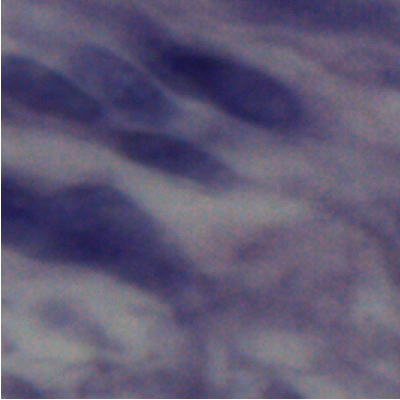} &
		\includegraphics[width=0.150\textwidth]{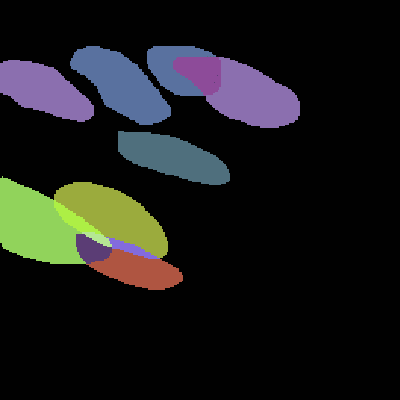}& 
		\includegraphics[width=0.150\textwidth]{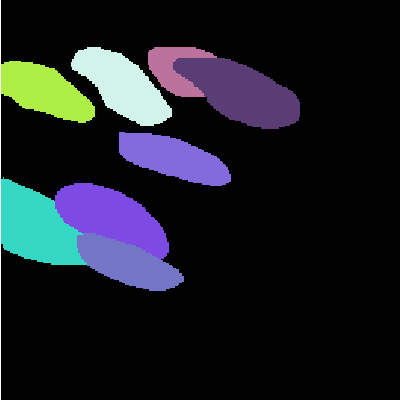} &
		\includegraphics[width=0.150\textwidth]{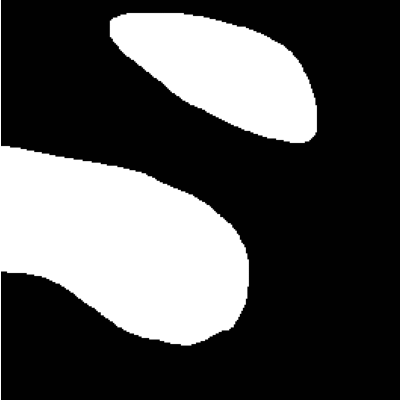} \\		
	\end{tabular}
	\caption{Example from the NuInsSeg dataset showing, from left to right: the H\&E-stained tissue image, the labeled mask with preserved overlapping areas, the labeled mask with overlapping regions assigned to the last encountered instance, and the corresponding vague area mask. }
	\label{fig:nuinsseg_example}
\end{figure}

To conduct our experiments, we used the NuInsSeg dataset~\citep{mahbod2024nuinsseg} to demonstrate the effect of our modifications as the main evaluation dataset. Although several publicly available datasets exist for this task~\citep{torbati2026nucfuserank}, we specifically selected the NuInsSeg dataset because it has unique characteristics that make it particularly suitable for this study. First, it is one of the largest publicly available and fully manually annotated datasets for nuclear instance segmentation of Hematoxylin and Eosin (H\&E)-stained histologic images, with 30,698 segmented nuclei across 665 image patches of size $512\times512$ pixels extracted from 31 organs. 
Moreover, NuInsSeg is the only available dataset that provides vague area annotations for the entire image, whereas other datasets do not include this additional annotation. Furthermore, NuInsSeg provides two manual mask types: one where overlapping areas are preserved, and another where the overlapping regions are assigned to the last encountered instance in the annotation file. Having both versions is important for investigating the effect of our proposed modification for handling overlapping areas. 
An example image with corresponding segmentation masks (with preserved overlapping areas and merged overlapping areas), along with corresponding vague area annotations, is shown in Figure~\ref{fig:nuinsseg_example}. 

Besides NuInsSeg, we used three additional datasets, namely CryoNuSeg \citep{mahbod2021cryonuseg_org}, PCNS~\citep{hou2020dataset}, and MoNuSAC~\citep{9446924}, for further experiments in this study, although the effect of all modifications could not be investigated with every dataset, as some required information (e.g., annotations of vague areas) was unavailable.

CryoNuSeg contains 30 H\&E-stained histologic image patches extracted from 10 human organs with a fixed size of $512\times512$ pixels. Similar to NuInsSeg, annotations for overlapping areas are provided in this dataset, but vague area annotations are not, so only 3 out of 4 of our modifications could be investigated. In addition, CryoNuSeg provides multiple manual annotations of the same images: the images were annotated by two annotators, and the first annotator repeated the annotation in a second round~\citep{mahbod2021cryonuseg_org}. We used these annotations to measure inter-observer agreement (annotator 1 vs.\ annotator 2) and intra-observer agreement (annotator 1, round 1 vs.\ round 2) under the standard and the modified evaluation pipelines.

PCNS is another benchmark dataset containing 1,356 H\&E-stained histologic image patches with a fixed size of $256\times256$ pixels. Annotations for vague areas and overlapping areas are not provided for this dataset, so only 2 out of 4 modifications in the evaluation pipeline could be investigated for this dataset.

MoNuSAC is a challenge dataset for nuclear instance segmentation and classification, containing H\&E-stained images from four organs (breast, kidney, lung, and prostate) with varying image sizes~\citep{9446924}. We used its test set, which consists of 101 images and includes annotations of ambiguous regions. Since overlapping regions are assigned to a single instance in the MoNuSAC annotations, Modification \#3 could not be investigated, and only Modifications \#1, \#2, and \#4 were applied. We used the publicly available predictions of four top-ranked challenge teams (denoted PL1, PL2, PL3, and L2) to investigate whether the proposed modifications change the ranking of methods in a realistic benchmarking setting. Since this study focuses on nuclear instance segmentation, and to be consistent with the other datasets used in this study, which do not provide class labels, all nucleus classes in MoNuSAC were merged into a single class.

Finally, we constructed a synthetic example with four circular instances, each designed to isolate the effect of a single modification, which is described in Section~\ref{sec:synthetic}.

\subsection{Incorporating vague areas into evaluation (Modification \#1)}
\label{sec:vague areas}

Most publicly available datasets for nuclear instance segmentation provide RGB tissue images, commonly extracted as patches from WSIs, along with corresponding labeled masks. In these labeled masks, each instance (nucleus) is typically assigned a unique integer identifier, while the background is represented by zero. These instance masks can be easily converted into binary masks for semantic segmentation analysis. While these two types of data (tissue images and labeled masks) are essential for both training and evaluation of models, most datasets do not provide annotations for ambiguous regions. Ambiguous regions refer to areas in the images where precise and reliable annotation is not feasible, even for human experts. These regions typically arise from factors such as out-of-focus areas, tissue folds or tears, densely packed nuclei with indistinguishable boundaries, or areas where stain quality is insufficient to clearly delineate individual nuclei. Ignoring such regions, particularly during evaluation, may lead to unfair assessment, as model predictions in these areas may be either penalized or rewarded in a manner that is not meaningful. Among the available datasets for nuclear instance segmentation, as presented in~\citep{torbati2026nucfuserank}, only the NuInsSeg dataset~\citep{mahbod2024nuinsseg} provides explicit annotations for ambiguous regions across entire images, making it particularly suitable for our study. The MoNuSAC dataset~\citep{9446924} also includes annotations of ambiguous regions in its test set; however, no nuclei are annotated within these regions, which prevents a full analysis of their effect on the ground truth. Nevertheless, these masks still allow predicted instances within ambiguous regions to be excluded from the evaluation, and we therefore applied this modification in our ranking analysis on MoNuSAC.

To account for ambiguous regions in the evaluation pipeline, we exclude them from both the model predictions and the ground truth segmentation masks. This avoids penalizing or rewarding predictions in regions that cannot be reliably annotated. However, a potential issue arises for instances (in both ground truth and predictions) that intersect with ambiguous regions (that is, instances that are partially inside and partially outside these regions). To address this, we investigated three different strategies: (i) removing all instances that have any overlap with ambiguous regions, (ii) removing only those instances that are entirely contained within ambiguous regions, and (iii) applying a controllable threshold-based strategy, which generalizes both previous cases. In the latter approach, which we implemented in the study, an instance is removed if the ratio between its overlap area with the ambiguous region and its total area exceeds a predefined threshold. 
We experimented with multiple threshold values and found the metrics to be relatively insensitive to this parameter (Table~\ref{tab:amb_threshold}); we therefore recommend a threshold of 0.25 as the default, as it achieved the best overall performance in our experiments. However, it should be noted that this parameter is an optional input parameter for the NucEval function (named \texttt{overlap\_thresh\_amb}) and can be changed based on the application. After removing the affected instances and zeroing out the ambiguous pixels, the remaining instances are relabeled to ensure contiguous integer identifiers for instances, which is required for metric computation. The pseudocode for this modification is provided in Algorithm~\ref{alg:ambiguous_evaluation}. 

\begin{algorithm}[]
	\caption{Ambiguous region handling (Modification \#1)}
	\label{alg:ambiguous_evaluation}
	\begin{algorithmic}[1]
		
		\State \textbf{Input:} Ground truth masks $G$, prediction $P$, ambiguous mask $A$, threshold $\tau$ ($\texttt{overlap\_thresh\_amb}$)
		\State \textbf{Output:} PQ, AJI, Dice, DQ, SQ
		
		\State $\tilde{G} \gets G$, $\tilde{P} \gets P$
		
		\For{each instance in $\tilde{G}$ and $\tilde{P}$}
		\State Compute overlap ratio with ambiguous regions
		\If{overlap ratio $> \tau$}
		\State Remove instance
		\EndIf
		\EndFor
		
		\State Relabel instances to contiguous IDs
		
		\State Compute metrics using $\tilde{G}, \tilde{P}$
		
		\State \textbf{return} metrics
		
	\end{algorithmic}
\end{algorithm}


\subsection{Score normalization (Modification \#2)}
\label{sec:score_normalization}
Another potential issue that might affect the evaluation of nuclear instance segmentation is the way scores are aggregated. In many studies and challenges (e.g., the MoNuSeg challenge~\citep{monuseg}), metrics are calculated for each image and then averaged over all obtained scores. However, histologic image patches might contain different numbers of nuclei. For patches with only a few nuclei, imperfect instance segmentation can disproportionately affect the overall score, introducing a bias toward images with few annotated nuclei. This does not reflect typical practice, as instance segmentation is usually performed on WSIs or large regions of interest rather than on individual patches. One way to address this issue is to merge all patches in the test set into a single large image and compute the metrics on that, as proposed for the tissue segmentation task in the PUMA challenge~\citep{schuiveling2025novel, Schuiveling2026.03.09.26347935}. However, this approach can be computationally expensive when dealing with a large number of images and limited computational resources. In this study, we instead adopt a nuclear count-normalized average as a practical alternative, where each image's contribution to the final score is weighted by the number of nuclei in its GT mask. For images whose GT mask contains no nuclei, the number of predicted instances is used as the weight instead, so false positive detections on such images are still penalized. This removes the hidden bias toward sparsely annotated patches. A pseudocode description of this normalization procedure is provided in Algorithm~\ref{alg:normalization}. 

\begin{algorithm}[]
	\caption{Nuclear count-normalized evaluation (Modification \#2)}
	\label{alg:normalization}
	\begin{algorithmic}[1]
		\State \textbf{Input:} Ground truth masks $\{G_1, \dots, G_N\}$, predictions $\{P_1, \dots, P_N\}$
		\State \textbf{Output:} Weighted metric scores
		\For{each image $I_i$}
		\State Compute metric score $s_i$ (PQ, AJI, Dice, DQ, SQ)
        \State Count number of GT nuclei $n_i$ in $G_i$; if $n_i = 0$, set $n_i$ to the number of predicted instances in $P_i$
		\EndFor
		\State Compute weighted average: $\bar{s} = \frac{\sum_{i=1}^{N} n_i \cdot s_i}{\sum_{i=1}^{N} n_i}$
		\State \textbf{return} normalized scores ($\bar{s}$)
	\end{algorithmic}
\end{algorithm}


\subsection{Handling of overlapping regions (Modification \#3)}
\label{sec:ovelapped}
Separating overlapping nuclei is one of the main challenges in instance segmentation, and various strategies such as watershed-based post-processing and distance-based models have been proposed to address it~\citep{graham2019hover, pmlr-v250-baumann24a, naylor2018segmentation}.
However, a critical issue remains in the evaluation pipeline. In many datasets, overlapping regions are arbitrarily assigned to one of the instances, which may unfairly penalize model predictions. For example, in the MoNuSeg and MoNuSAC challenges, overlapping regions are assigned to the last encountered instance in the annotation file~\citep{monuseg, 9446924, FOUCART2023109600}. This means that the order of nuclear annotations can affect the evaluation outcome, which is not logically justified.
To address this problem, we propose assigning overlapping regions to all relevant instances instead of arbitrarily assigning them to a single instance (last encountered instance). This approach preserves the information in the original ground-truth segmentation mask and avoids unnecessary penalization of model predictions. To assign overlapping regions to all instances that share them, the segmentation mask can be represented as a list of binary masks, where each instance is stored as a separate binary mask while preserving the overlapping regions. As a result, the overlapping region appears in all instances that share that area. It should be noted that such information cannot be extracted directly from labeled masks, as each pixel can only hold a single integer label. However, some datasets such as NuInsSeg provide sets of regions of interest (ROIs) for each instance in each image, from which individual binary masks can be derived and used to compute the metrics appropriately. In the NucEval function, the ground truth can be provided in three formats: as a set of ROI files, as a list of binary masks where each mask represents one instance, or as a labeled mask. The latter format is used when overlapping information is not available, but the function can still operate. 
An illustrative synthetic example of this modification, in which the shared region of two overlapping instances is assigned to opposite instances in the ground truth and the prediction, is provided in Section~\ref{sec:synthetic}.
A pseudocode description of this modification is provided in Algorithm~\ref{alg:overlap}.

\begin{algorithm}[]
	\caption{Overlapping region handling (Modification \#3)}
	\label{alg:overlap}
	\begin{algorithmic}[1]
		\State \textbf{Input:} Ground truth $G$ (ROI set, list of binary masks, or label map), prediction $P$
		\State \textbf{Output:} PQ, AJI, Dice, DQ, SQ
		\If{$G$ is a set of ROI files}
		\State Convert each ROI polygon to a binary mask
		\State $\mathcal{M}_G \gets \{m_1, m_2, \dots, m_K\}$ \
		\ElsIf{$G$ is a list of binary masks}
		\State $\mathcal{M}_G \gets G$ 
		\ElsIf{$G$ is a label map}
		\State Extract each instance as a binary mask
		\State $\mathcal{M}_G \gets \{m_1, m_2, \dots, m_K\}$ 
		\EndIf
		\State Compute metrics using $\mathcal{M}_G$ and $P$
		\State \textbf{return} metrics
	\end{algorithmic}
\end{algorithm}


\subsection{Border delineation uncertainty handling (Modification \#4)}
\label{sec:border_uncertain}

Annotation uncertainty is one of the common issues impacting manual annotation. Annotation uncertainty, particularly at object boundaries, has previously been reported in radiologic images~\citep{10.5555/3495724.3497045}. Variations in boundary delineation have also been shown to affect radiomic features and classical machine learning predictions in radiology~\citep{HATAMIKIA202452}. This issue can be even more severe in the case of nuclear segmentation, as a single image patch may contain hundreds of nuclei, each of which can be affected by boundary delineation uncertainty~\citep{FOUCART2023109600, HARTMAN20207}. Consequently, penalizing a model for disagreeing with the ground truth by a few pixels at the boundary is arguably unfair, as even human annotators would likely disagree on the exact boundary location.



To address this, we propose creating a boundary uncertainty zone around each instance and excluding this zone from the evaluation. The zone is constructed by applying morphologic erosion and dilation to each instance mask independently. The zone is defined as the difference between the dilated and eroded masks, thereby capturing the zone of uncertainty in both directions (inward and outward) from the annotated boundary. The per-instance zones are then combined into a single global zone mask through a union operation, and all pixels within this zone are zeroed out from both the ground truth and the prediction before computing any metrics. The width of the zone is controlled by the \texttt{zone\_width} parameter in the NucEval function, which specifies the number of pixels used for both erosion and dilation. Setting \texttt{zone\_width} to 0 disables this modification entirely. 

It is important to note that metric scores generally increase with the zone width (Table~\ref{tab:ring_size}), and excessively large values can eliminate small nuclei entirely and reduce the evaluation to only the confident core regions of larger nuclei. We therefore use the minimum meaningful width of one pixel as a conservative default in all our experiments. This choice reflects the fact that nuclear boundaries are inherently difficult to delineate and that manual annotations of the same nuclei commonly differ at their boundaries, while keeping the excluded region as small as possible. The zone width is a user-defined parameter in NucEval (set to one pixel by default and in all our experiments) and can be increased for datasets with larger nuclei or images acquired at a higher magnification; however, when NucEval is used for benchmarking, the zone width should be fixed a priori and applied identically to all compared methods, so that it cannot be tuned to inflate the scores of a specific method.
A pseudocode description of this procedure is provided in Algorithm~\ref{alg:ring}, and example images illustrating the effect of zone size on the labeled segmentation masks are shown in Figure~\ref{fig:example_ring} in the appendix.   

\begin{algorithm}[]
	\caption{Evaluation with border uncertainty zone (Modification \#4)}
	\label{alg:ring}
	\begin{algorithmic}[1]
		\State \textbf{Input:} Ground truth masks $\{G_1, \dots, G_N\}$, prediction $P$, zone width $r$ (\texttt{zone\_width})
		\State \textbf{Output:} PQ, AJI, Dice, DQ, SQ
		\State Initialize zone mask $R \gets \mathbf{0}$
		\For{each instance mask $m_i \in G$}
		\State $m_i^{+} \gets \text{dilate}(m_i, r)$ 
		\State $m_i^{-} \gets \text{erode}(m_i, r)$ 
		\State $R \gets R \cup (m_i^{+} - m_i^{-})$ 
		\EndFor
		\For{each instance mask $m_i \in G$}
		\State Zero out zone pixels in $m_i$
		\EndFor
		\State Zero out zone pixels in $P$
		\State Relabel instances to contiguous IDs
		\State Compute metrics using modified masks and predictions
		\State \textbf{return} metrics
	\end{algorithmic}
\end{algorithm}

\subsection{Overall NucEval function}

Considering all modifications described above, we propose the NucEval function that can incorporate the proposed modifications into the evaluation pipeline. It should be noted that NucEval is modular and can operate without any modifications (standard metrics), with all modifications enabled, or with any selected subset of modifications. Modifications \#2 and \#4 require only the standard labeled masks and can therefore be applied to any dataset, whereas Modifications \#1 and \#3 depend on additional annotations/information, namely vague area masks and overlap-preserving instance annotations (ROI sets or per-instance binary masks), respectively. 


\subsection{Deep Learning-based models}
For the evaluation analysis, we used three state-of-the-art deep learning (DL)-based models in this study. Specifically, we used HoVer-Net~\citep{graham2019hover}, HoVer-NeXt~\citep{pmlr-v250-baumann24a}, and CellViT~\citep{HORST2024103143} for training, evaluation and reporting of results. All three models are benchmark methods and collectively represent both major DL paradigms for image analysis, i.e., CNN-based and ViT-based architectures, which allows us to show that the effect of the proposed modifications is model-agnostic. These models follow an encoder-multi-decoder design, where the outputs from different decoders are merged to produce the final instance segmentation masks. Further details about these models can be found in their corresponding publications. For the MoNuSAC dataset, we did not train any models; instead, we directly evaluated the publicly available predictions of the four top-ranked challenge teams.

\subsection{Implementation details}
\label{sec:implementation_details}
The NucEval implementation is based on Python and requires \texttt{numpy}, \texttt{scipy}, and \texttt{opencv-python} to operate. The standard version of the metrics, without modification, is based on the code provided in the HoVer-Net implementation~\citep{graham2019hover}. The main parameters of the evaluation are exposed as user-defined inputs with default values: the ambiguous-region overlap threshold (\texttt{overlap\_thresh\_amb}, 0.25 by default), the border uncertainty zone width (\texttt{zone\_width}, one pixel by default), and the IoU threshold for instance matching (\texttt{match\_iou}, 0.5 by default). For reporting the results, we used the default values for all three parameters. The sensitivity of the evaluation to these parameters is reported in \ref{sec:hyperparameter}, in Table~\ref{tab:amb_threshold} for \texttt{overlap\_thresh\_amb}, Table~\ref{tab:ring_size} for \texttt{zone\_width}, and Table~\ref{tab:iou_sweep} for \texttt{match\_iou} (from 0.5 to 0.9). In addition to the normalized scores, NucEval also reports the corresponding non-normalized scores. The training and evaluation pipelines of the utilized DL models (HoVer-Net, HoVer-NeXt, and CellViT) are based on the original architectures and implementations from their respective studies. We trained the models using 5-fold cross-validation with a fixed seed to ensure reproducible data splits. Each model was trained for 150 epochs in each cross-validation fold with a batch size of 4. Training was performed on a single workstation equipped with an Intel Core i9-14900KF processor, 64~GB of RAM, and a single NVIDIA GeForce RTX 4090 GPU using the PyTorch deep learning framework.
Our full implementation, along with guidelines and examples for both single-image and dataset-level evaluation, as well as the training code for the DL models, is publicly available at: \url{https://github.com/masih4/nuc_eval}.


\section{Results \& Discussion}
\label{sec:res}
\subsection{Impact of individual modifications}

The results for the impact of each modification on the segmentation metrics compared to the baseline (no modification) based on three utilized models (HoVer-Net, HoVer-NeXt, and CellViT) for the entire NuInsSeg dataset are shown in Table~\ref{tab:nuinsseg_entire}. These show that, in almost all cases, applying the modifications yields increased metrics across all three models. Modification \#4 (border uncertainty zone) shows the most substantial impact, with PQ improvements of approximately 5-6 percentage points across all models compared to the baseline. 

Modification \#1 (ambiguous region handling) consistently improves all metrics, while Modification \#2 (score normalization) and Modification \#3 (overlapping region handling) yield improvements in most cases. These trends are consistent across all three models, demonstrating that the proposed modifications are model-agnostic and generally applicable.


\begin{table}[]
	\centering
	\caption{Effect of independently applied evaluation protocol modifications 
		on the entire NuInsSeg dataset (baseline vs.\ modified\#1-\#4).
		PQ: Panoptic Quality; AJI: Aggregated Jaccard Index;
		DQ: Detection Quality; SQ: Segmentation Quality.
		All values are reported in \%.}
	\begin{tabular}{ll ccccc}
		\hline
		\textbf{Model} & \textbf{Variant} & \textbf{PQ} & \textbf{AJI} & \textbf{DICE} & \textbf{DQ} & \textbf{SQ} \\
		\hline
		\multirow{5}{*}{\textbf{HoVer-Net}}
		& Baseline    & 53.40 & 57.50 & 80.29 & 70.38 & 75.17 \\
		& Modified \#1 & 56.17 & 60.75 & 81.31 & 73.28 & 75.82 \\
		& Modified \#2 & 54.42 & 56.74 & 83.22 & 71.37 & 75.95 \\
		& Modified \#3 & 54.65 & 58.53 & 80.35 & 72.85 & 74.37 \\
		& Modified \#4 & 59.04 & 61.20 & 82.93 & 72.82 & 80.34 \\
		\hline
		\multirow{5}{*}{\textbf{HoVer-NeXt}}
		& Baseline    & 54.17 & 58.45 & 82.59 & 72.14 & 74.40 \\
		& Modified \#1 & 56.80 & 61.06 & 83.81 & 74.83 & 75.17 \\
		& Modified \#2 & 55.96 & 59.44 & 85.19 & 74.18 & 75.14 \\
		& Modified \#3 & 54.70 & 58.85 & 82.70 & 73.18 & 74.14 \\
		& Modified \#4 & 60.01 & 62.37 & 85.46 & 74.58 & 79.78 \\
		\hline
		\multirow{5}{*}{\textbf{CellViT}}
		& Baseline    & 56.44 & 60.03 & 81.92 & 74.50 & 74.83 \\
		& Modified \#1 & 59.72 & 63.31 & 83.17 & 77.95 & 75.58 \\
		& Modified \#2 & 58.33 & 60.54 & 84.80 & 76.68 & 75.79 \\
		& Modified \#3 & 57.46 & 60.89 & 81.86 & 76.18 & 74.60 \\
		& Modified \#4 & 62.69 & 64.32 & 84.97 & 77.23 & 80.41 \\
		\hline
	\end{tabular}
	\label{tab:nuinsseg_entire}
\end{table}

The impact of Modification \#2 (score normalization) is inherently more pronounced for datasets with imbalanced images in terms of the number of nuclei per image. To investigate this, we extracted a subset of NuInsSeg using the following criteria: PQ greater than 70\% and number of nuclei greater than 80, or PQ less than 30\% and number of nuclei less than 15 (based on HoVer-NeXt results). The distributions of PQ scores for the entire NuInsSeg dataset and the extracted subset are shown in Figure~\ref{fig:scaterplot} in the appendix, and the full results for the derived subset dataset are reported in Table~\ref{tab:nuinsseg_subset}. 

\begin{table}[H]
	\centering
	\caption{Effect of evaluation protocol modifications on the subset of NuInsSeg
		dataset (baseline vs.\ modified\#1-\#4).
		PQ: Panoptic Quality; AJI: Aggregated Jaccard Index;
		DQ: Detection Quality; SQ: Segmentation Quality.
		All values are reported in \%.}
	\begin{tabular}{ll ccccc}
		\hline
		\textbf{Model} & \textbf{Variant} & \textbf{PQ} & \textbf{AJI} & \textbf{DICE} & \textbf{DQ} & \textbf{SQ} \\
		\hline
		\multirow{5}{*}{\textbf{HoVer-Net}}
		& Baseline     & 28.68 & 34.42 & 54.65 & 38.79 & 62.79 \\
		& Modified \#1 & 40.03 & 46.92 & 63.74 & 49.84 & 65.65 \\
		& Modified \#2 & 55.38 & 57.14 & 80.53 & 68.66 & 78.26 \\
		& Modified \#3 & 29.89 & 35.40 & 54.74 & 41.12 & 61.26 \\
		& Modified \#4 & 33.23 & 36.28 & 56.16 & 42.32 & 66.58 \\
		\hline
		\multirow{5}{*}{\textbf{HoVer-NeXt}}
		& Baseline     & 28.00 & 36.77 & 59.01 & 38.06 & 64.87 \\
		& Modified \#1 & 40.35 & 49.15 & 69.98 & 51.14 & 67.36 \\
		& Modified \#2 & 60.14 & 63.05 & 83.41 & 74.56 & 78.38 \\
		& Modified \#3 & 28.20 & 37.46 & 59.32 & 38.41 & 65.06 \\
		& Modified \#4 & 30.96 & 38.65 & 60.30 & 38.82 & 70.34 \\ 
		\hline
		\multirow{5}{*}{\textbf{CellViT}}
		& Baseline     & 29.07 & 36.93 & 56.30 & 38.57 & 63.44 \\
		& Modified \#1 & 43.45 & 51.78 & 66.72 & 53.46 & 66.42 \\
		& Modified \#2 & 60.27 & 62.76 & 82.64 & 74.73 & 78.81 \\
		& Modified \#3 & 30.10 & 37.61 & 56.20 & 41.09 & 64.89 \\
		& Modified \#4 & 33.47 & 39.56 & 58.28 & 41.81 & 71.32 \\
		\hline
	\end{tabular}
	\label{tab:nuinsseg_subset}
\end{table}

The results show that the impact of Modification \#2 is much more evident for this subset, with PQ improvements of approximately 27-32 percentage points compared to the baseline. Interestingly, for Modification \#1, the results are also substantially improved, with PQ gains of approximately 11-14 percentage points.

The results for both the full NuInsSeg dataset (Table~\ref{tab:nuinsseg_entire}) and the NuInsSeg subset (Table~\ref{tab:nuinsseg_subset}) show that Modification \#4 generally has a consistent positive impact, while the effects of Modifications \#1 and \#2 are dataset-dependent and vary based on the number of nuclei and vague area annotations present across the dataset.

Modification \#3 in both cases shows the least impact on the results, although it leads to improved performance in most cases. As overlapping areas usually constitute a small portion of the dataset, this modification may have a limited effect on the overall metrics. To demonstrate the maximum potential impact of Modification \#3, we computed the scores using two formats for the provided annotation masks in the NuInsSeg dataset: ground truth labeled masks where overlapping areas are merged to one of the instances, versus ground truth with ROI sets where overlapping areas are preserved. This results in PQ, AJI, Dice, DQ, and SQ of 88.20\%, 89.66\%, 96.02\%, 99.35\%, and 88.87\%, respectively. In an ideal case, the results should be perfect (100\%), and the inferior performance is due to the merging of overlapping areas, which demonstrates that this modification can affect the results by a notable margin. The effect of Modification \#3 in a controlled setting is further illustrated with a synthetic example in Section~\ref{sec:synthetic}.

\subsection{Impact of cumulative modifications}
\begin{figure}
	\centering
	\includegraphics[width=0.9\linewidth]{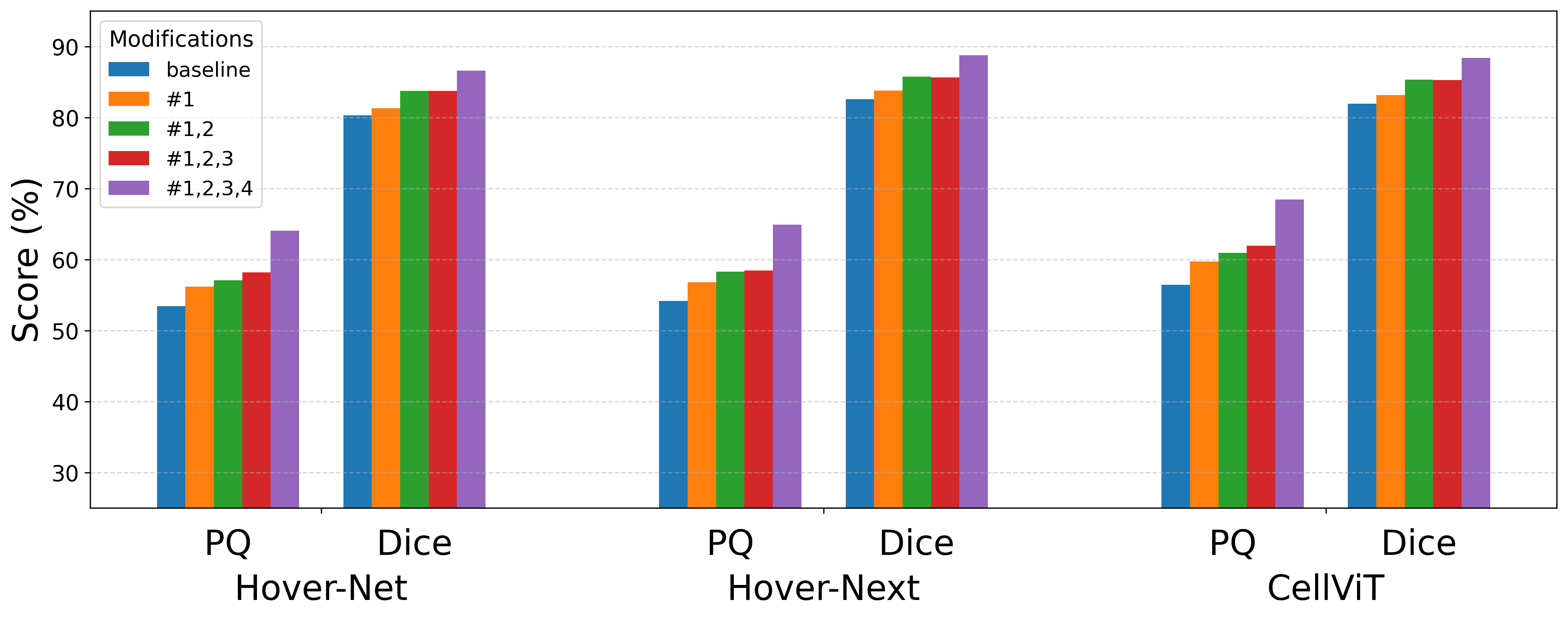}
	\caption{Panoptic Quality (PQ) and Dice score comparison across models and modifications on the entire NuInsSeg dataset. For each model, the results are shown separately for PQ and Dice. All modifications are applicable to this dataset. }
	\label{fig:res_pq_entire}
\end{figure}

\begin{figure}
	\centering
	\includegraphics[width=0.9\linewidth]{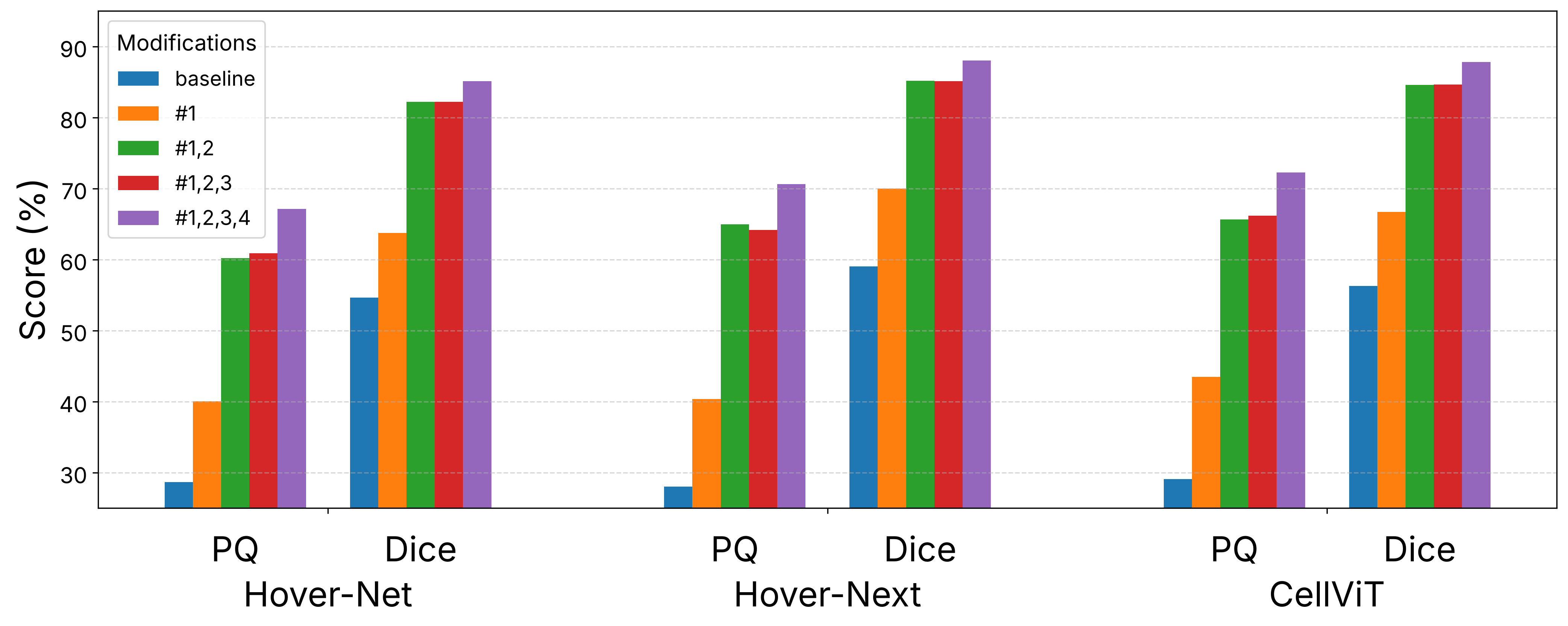}
	\caption{Panoptic Quality (PQ) and Dice score comparison across models and modifications on the subset of the NuInsSeg dataset. For each model, the results are shown separately for PQ and Dice. All modifications are applicable to this dataset.}
	\label{fig:res_pq_subset}
\end{figure}

In the previous section, we investigated the impact of each modification on
the results independently (i.e., only one modification was applied at a time).
However, these modifications can also be applied simultaneously. To show
the cumulative impact of the modifications on the results, we present the
PQ and Dice scores for all three models for the entire NuInsSeg dataset and
the extracted subset of NuInsSeg in Figure~\ref{fig:res_pq_entire} and 
Figure~\ref{fig:res_pq_subset}, respectively. We also provide the complete 
results for all metrics for both cases in Table~\ref{tab:nuinsseg_entire_cumulative} 
and Table~\ref{tab:nuinsseg_subset_cumulative} in the appendix. As observed, 
applying all modifications in a cumulative manner increases PQ by 10.67, 10.75, 
and 12.00 percentage points for HoVer-Net, HoVer-NeXt, and CellViT, respectively, compared to the baseline on the entire dataset. Similarly, Dice scores improve 
by 6.31, 6.17, and 6.48 percentage points for the three models, respectively.

The same trend can also be observed for the subset dataset, where the impact is more evident, with PQ increases of 38.45, 42.61, and 43.19 percentage points for HoVer-Net, HoVer-NeXt, and CellViT, respectively. 
The Dice scores also show a more pronounced improvement on the subset,  with gains of 30.45, 29.03, and 31.51 percentage points, respectively,  highlighting the substantial effect of the proposed modifications on imbalanced datasets.

\begin{figure}
	\centering
	\includegraphics[width=0.9\linewidth]{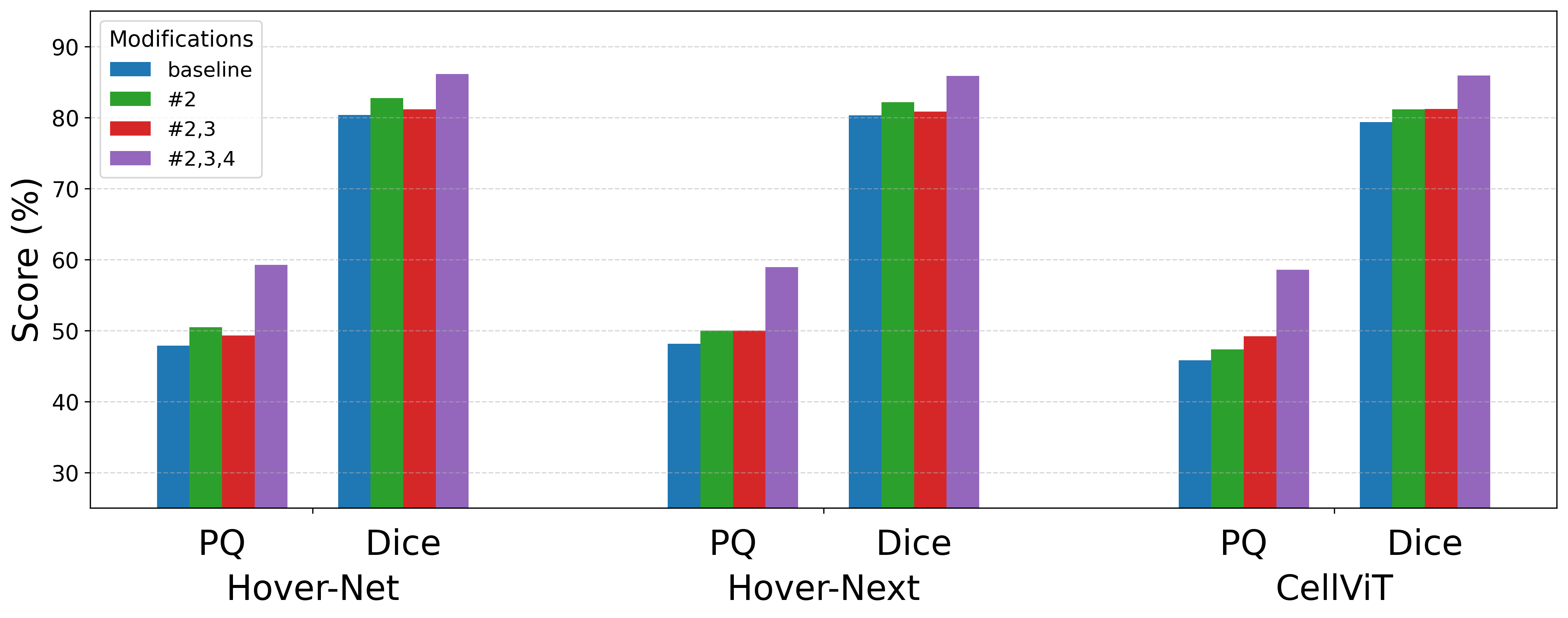}
	\caption{Panoptic Quality (PQ) and Dice score comparison across models and modifications on the CryoNuSeg dataset. For each model, results are shown separately for PQ and Dice. Modification \#1 is not applicable to this dataset.}
	\label{fig:res_pq_cryonuseg}
\end{figure}

\begin{figure}
	\centering
	\includegraphics[width=0.9\linewidth]{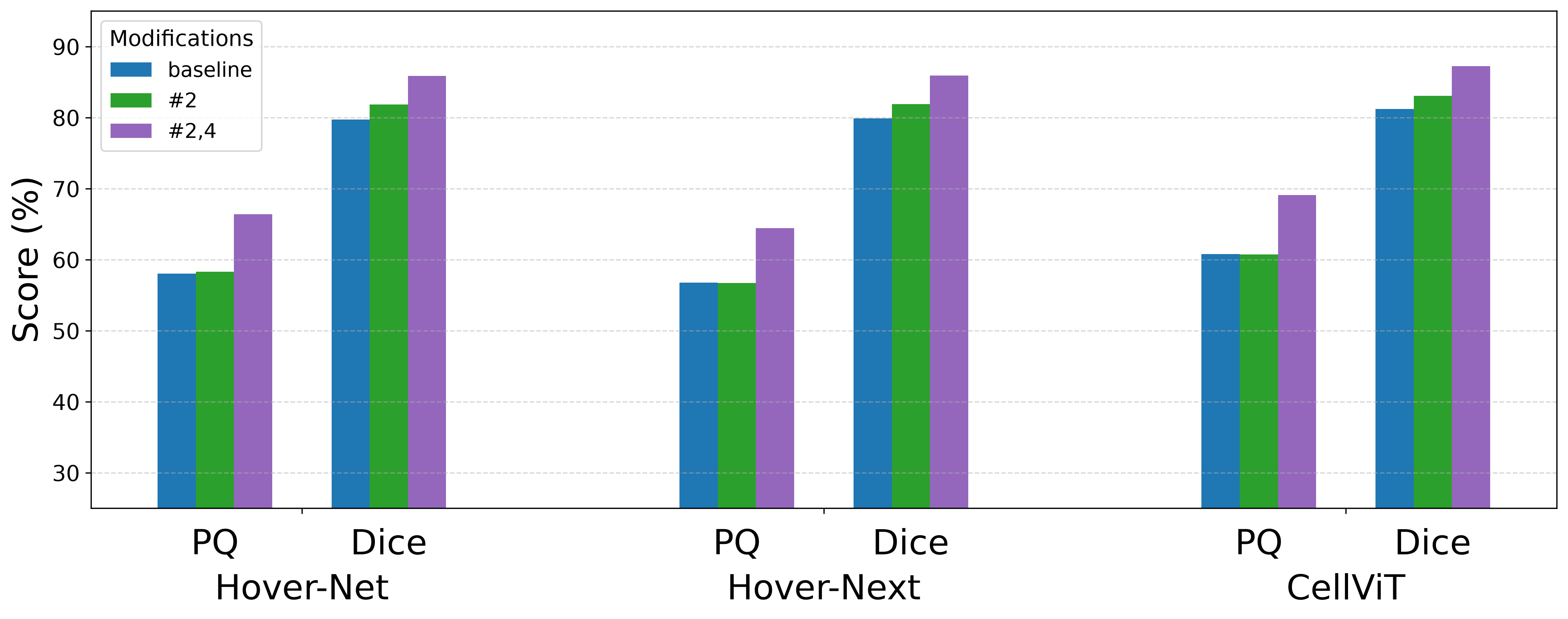}
	\caption{Panoptic Quality (PQ) and Dice score comparison across models and modifications on the PCNS dataset. For each model, the results are shown separately for PQ and Dice. Modifications \#1 and \#3 are not applicable to this dataset.}
	\label{fig:res_pq_pcns}
\end{figure}

In addition to the NuInsSeg dataset, we also report the cumulative results based on the applied modifications for the CryoNuSeg and PCNS datasets in Figure~\ref{fig:res_pq_cryonuseg} and Figure~\ref{fig:res_pq_pcns}, respectively, based on PQ and Dice scores. The complete results for all metrics are reported in Table~\ref{tab:cryonuseg_entire_cumulative} and Table~\ref{tab:pcns_entire_cumulative} in the appendix. It should be noted that Modification \#1 is not applicable to CryoNuSeg, as vague area annotations are not available, while both Modifications \#1 and \#3 are not applicable to PCNS, as neither vague area annotations nor ROI annotations for individual instances are provided in this dataset.

For the CryoNuSeg dataset, applying all available modifications cumulatively increases PQ by 11.38, 10.80, and 12.76 percentage points for HoVer-Net, HoVer-NeXt, and CellViT, respectively, compared to the baseline. Dice scores also improve by 5.76, 5.54, and 6.58 percentage points for the three models, respectively.
These gains are consistent with the trends observed on the NuInsSeg dataset, further supporting the generalizability of the proposed modifications. Regarding the individual contributions, Modification \#2 (score normalization) improves both PQ and Dice across all three models, with PQ gains of 2.63, 1.84, and 1.56 percentage points and Dice gains of 2.41, 1.85, and 1.80 percentage points for HoVer-Net, HoVer-NeXt, and CellViT, respectively. The effect of Modification \#3 (overlapping region handling), however, varies across models. CellViT benefits from this modification, with PQ increasing from 47.35\% to 49.21\% when adding overlap handling, while Dice remains stable (81.15\% to 81.20\%). HoVer-Net shows a decline in both PQ (50.47\% to 49.31\%) and Dice (82.74\% to 81.15\%), and HoVer-NeXt exhibits stable PQ (49.96\% to 49.97\%) but decreased Dice (82.16\% to 80.85\%). This is consistent with the behavior observed on the NuInsSeg dataset, where the effect of Modification \#3 is less evident compared to other modifications.

For the PCNS dataset, where only Modifications \#2 and \#4 could be applied, PQ still increases by 8.33, 7.66, and 8.35 percentage points for HoVer-Net, HoVer-NeXt, and CellViT, respectively. Dice scores improve by 6.16, 6.01, and 6.02 percentage points, respectively. 
The majority of the PQ improvement on PCNS comes from Modification \#4
(border uncertainty zone), as Modification \#2 (score normalization) alone
has almost no effect on PQ (e.g., PQ changes by only +0.23, $-$0.05, and
$-$0.04 percentage points for HoVer-Net, HoVer-NeXt, and CellViT,
respectively). However, it is worth noting that, while score normalization
has a negligible effect on PQ, it consistently improves Dice scores across
all three models, with gains of 2.11, 2.02, and 1.83 percentage points for
HoVer-Net, HoVer-NeXt, and CellViT, respectively. As shown in Figure~\ref{fig:pq_aji_vs_nuclei_PCNS} in the appendix, although the PCNS dataset contains images with varying nucleus counts, the majority of images are concentrated in the 10-50 nucleus range, with relatively consistent PQ scores in this region. Like the full NuInsSeg dataset, the PCNS dataset has a more balanced distribution than the NuInsSeg subset. Consequently, the effect of Modification \#2 (score normalization) is less evident here.

\subsection{Inter- and intra-observer agreement}
\label{sec:agreement}
To assess whether the proposed modifications bring the evaluation closer to expert consensus, we measured inter- and intra-observer agreement on the CryoNuSeg dataset, which provides multiple manual annotations of the same images. In this setting, one annotation is used as the reference and the other as the prediction. Ideally, two annotations of the same image by experts should yield perfect agreement, and any disagreement reflects both genuine annotation variability and artifacts introduced by the evaluation pipeline. If a modification reduces such artifacts, the agreement scores are expected to increase. Since vague area annotations are not available for CryoNuSeg, Modification \#1 could not be applied, and Modifications \#2, \#3, and \#4 were applied cumulatively. The results are reported in Table~\ref{tab:agreement}.

\begin{table}[h]
	\centering
	\caption{Intra-observer (annotator 1, round 1 vs.\ round 2) and inter-observer (annotator 1 vs.\ annotator 2) agreement on the CryoNuSeg dataset under the baseline and cumulatively modified evaluation pipelines. Modification \#1 is not applicable to CryoNuSeg, as vague area annotations are not available for this dataset.
		PQ: Panoptic Quality; AJI: Aggregated Jaccard Index;
		DQ: Detection Quality; SQ: Segmentation Quality.
		All values are reported in \%.}
	\begin{tabular}{ll ccccc}
		\hline
		\textbf{Agreement} & \textbf{Modifications} & \textbf{PQ} & \textbf{AJI} & \textbf{DICE} & \textbf{DQ} & \textbf{SQ} \\
		\hline
		\multirow{4}{*}{\textbf{Intra-observer}}
		& Baseline   & 56.92 & 62.30 & 83.75 & 76.31 & 74.44 \\
		& \#2        & 58.12 & 63.51 & 84.90 & 77.41 & 74.92 \\
		& \#2,3      & 58.61 & 63.40 & 85.53 & 78.33 & 74.69 \\
		& \#2,3,4    & 70.47 & 71.91 & 90.78 & 82.63 & 85.18 \\
		\hline
		\multirow{4}{*}{\textbf{Inter-observer}}
		& Baseline   & 50.78 & 54.06 & 78.89 & 68.48 & 73.87 \\
		& \#2        & 53.53 & 56.96 & 81.63 & 71.65 & 74.41 \\
		& \#2,3      & 54.69 & 57.54 & 82.60 & 72.84 & 74.80 \\
		& \#2,3,4    & 63.87 & 63.62 & 86.09 & 76.01 & 83.69 \\
		\hline
	\end{tabular}
	\label{tab:agreement}
\end{table}

As shown in Table~\ref{tab:agreement}, applying the modifications increases both intra- and inter-observer agreement compared to the baseline evaluation pipeline. With Modifications \#2, \#3, and \#4 applied, intra-observer PQ increases from 56.92\% to 70.47\% (13.55 percentage points) and inter-observer PQ increases from 50.78\% to 63.87\% (13.09 percentage points). Similar trends are observed for AJI and Dice, with gains of 9.61 and 7.03 percentage points for intra-observer agreement and 9.56 and 7.20 percentage points for inter-observer agreement, respectively. Score normalization (Modification \#2) and overlap handling (Modification \#3) lead to smaller increases, mainly in PQ, Dice, and DQ, whereas the largest increase is obtained with the border uncertainty zone (Modification \#4). These results indicate that the modified evaluation reduces disagreement stemming from annotation and evaluation artifacts rather than actual differences in the identified nuclei.

\subsection{Effect on model ranking}
\label{sec:ranking}
Beyond changes in absolute scores, we investigated whether the proposed modifications can change the ranking of methods. In our main experiments, the ranking of the three models is preserved on the entire NuInsSeg and PCNS, whereas on CryoNuSeg the two top-performing models exchange positions, with HoVer-NeXt achieving a slightly higher PQ than HoVer-Net under the baseline evaluation (48.12\% vs.\ 47.84\%) and HoVer-Net outperforming HoVer-NeXt after applying all available modifications (59.22\% vs.\ 58.92\%) (Table~\ref{tab:cryonuseg_entire_cumulative}). To assess this effect in a realistic benchmark setting, we re-evaluated the publicly available predictions of four top-ranked teams of the MoNuSAC challenge (denoted PL1, PL2, PL3, and L2) on the MoNuSAC test set. Since our study focuses on instance segmentation, all nucleus classes were merged into a single class, and the evaluation was performed in a class-agnostic manner. Moreover, since overlapping regions are assigned to a single instance in the MoNuSAC annotations, Modification \#3 could not be applied, and Modifications \#1, \#2, and \#4 were applied cumulatively. The PQ scores are reported in Table~\ref{tab:monusac}.

\begin{table}[h]
	\centering
	\caption{Panoptic quality scores of four top-ranked teams of the MoNuSAC challenge (PL1, PL2, PL3, and L2) on the MoNuSAC test set under the baseline and cumulatively modified evaluation pipelines. All nucleus classes are merged into a single class. Modification \#3 is not applicable to MoNuSAC, as overlapping regions are assigned to a single instance in its annotations. All values are reported in \%.}
	\begin{tabular}{l cccc}
		\hline
		\textbf{Modifications} & \textbf{PL1} & \textbf{PL2} & \textbf{PL3} & \textbf{L2} \\
		\hline
		Baseline   & 70.71 & 70.15 & 61.69 & 72.34 \\
		\#1        & 70.71 & 70.20 & 68.15 & 72.63 \\
		\#1,2      & 70.80 & 72.29 & 67.41 & 72.67 \\
		\#1,2,4    & 81.10 & 82.55 & 76.96 & 82.60 \\
		\hline
	\end{tabular}
	\label{tab:monusac}
\end{table}

Under the baseline evaluation, the ranking of the four teams is L2, PL1, PL2, and PL3. After applying Modifications \#1, \#2, and \#4, the ranking becomes L2, PL2, PL1, and PL3, with PL2 overtaking PL1 (82.55\% vs.\ 81.10\%). This change already occurs with Modifications \#1 and \#2 alone, where PL2 outperforms PL1 (72.29\% vs.\ 70.80\%). Although the top position of L2 is preserved, its margin over PL2 decreases from 2.19 to 0.05 percentage points. Moreover, PL3 shows the largest improvement, with its PQ increasing from 61.69\% to 76.96\%. Modification \#4 increases the PQ of all four teams by a similar amount (about 10 percentage points), whereas Modification \#1 mainly benefits PL3 (from 61.69\% to 68.15\%), which narrows its gap to the other teams. These results show that, among closely competing methods, the choice of evaluation pipeline can affect not only the absolute scores, but also model rank, and therefore, model selection, for a given task.

\subsection{Synthetic example}
\label{sec:synthetic}
\begin{figure}[h]
	\centering
	\includegraphics[width=\linewidth]{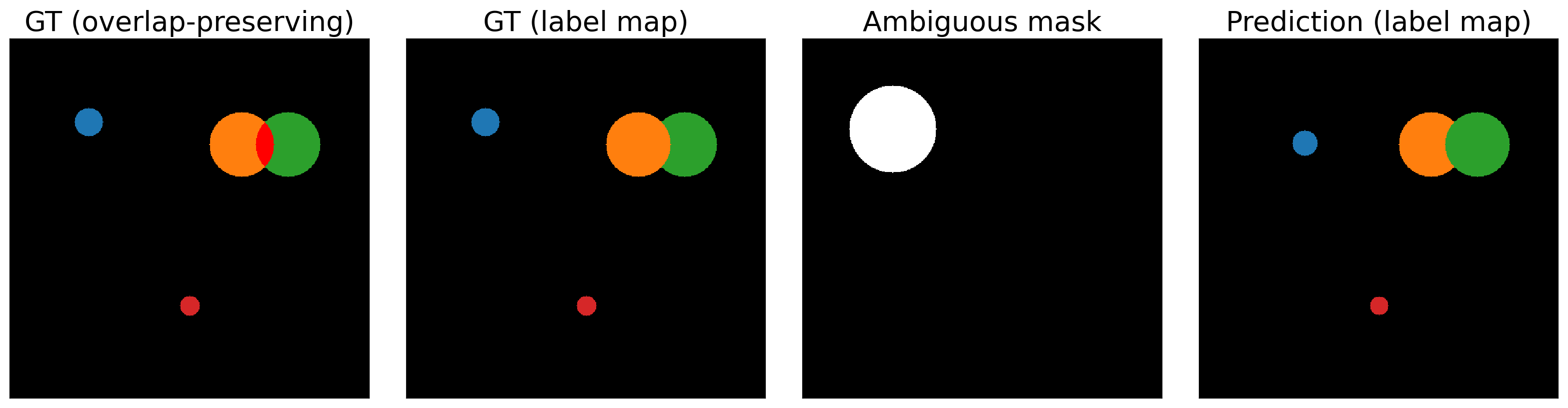}
	\caption{Synthetic example illustrating the effect of the proposed modifications. From left to right: the ground truth in the overlap-preserving format, where the region shared by the two overlapping instances is highlighted in red; the ground truth as a standard label map, where the overlapping region is merged into a single instance; the ambiguous-region mask; and the prediction label map. The example contains four circular instances, each designed to isolate one modification: instance 1 (top left) lies fully within the ambiguous region, where the ground truth and prediction do not overlap, illustrating Modification \#1; instances 2 and 3 (top right) overlap, with the shared region assigned to opposite instances in the ground truth and the prediction, illustrating Modification \#3; and instance 4 (bottom), a small circle whose prediction differs from the ground truth by a one-pixel boundary, illustrating Modification \#4.}
	\label{fig:synthetic}
\end{figure}

To illustrate the effect of each modification in a controlled setting, we constructed a synthetic example containing four circular instances, each designed to isolate a single modification, as shown in Figure~\ref{fig:synthetic}. Instance 1 (blue) lies fully within an ambiguous region, where the ground truth and prediction do not overlap (Modification \#1). Instances 2 and 3 (orange and green) overlap, and the shared region is assigned to opposite instances in the ground truth and the prediction (Modification \#3). Instance 4 (red) is a small circle whose prediction differs from the ground truth by a one-pixel boundary (Modification \#4). Modification \#2 (score normalization) is not included, as it is an aggregation step across images and is not applicable to a single image. The resulting scores are reported in Table~\ref{tab:synthetic}.

\begin{table}[h]
	\centering
	\caption{Scores for the synthetic example under the baseline and modified evaluation pipelines. Modification \#2 is not applicable to a single image.
		PQ: Panoptic Quality; AJI: Aggregated Jaccard Index;
		DQ: Detection Quality; SQ: Segmentation Quality.
		All values are reported in \%.}
	\begin{tabular}{l ccccc}
		\hline
		\textbf{Modifications} & \textbf{PQ} & \textbf{AJI} & \textbf{DICE} & \textbf{DQ} & \textbf{SQ} \\
		\hline
		Baseline   & 63.33 & 71.41 & 91.53  & 75.00  & 84.45 \\
		\#1        & 84.45 & 83.08 & 99.70  & 100.00 & 84.45 \\
		\#3        & 67.62 & 78.44 & 91.53  & 75.00  & 90.16 \\
		\#4        & 66.55 & 72.04 & 91.85  & 75.00  & 88.74 \\
		\#1,3      & 90.16 & 91.26 & 99.70  & 100.00 & 90.16 \\
		\#1,3,4    & 94.61 & 92.25 & 100.00 & 100.00 & 94.61 \\
		\hline
	\end{tabular}
	\label{tab:synthetic}
\end{table}

As shown in Table~\ref{tab:synthetic}, each modification removes the penalty associated with the specific artifact it targets. Modification \#1 removes the detection error caused by the instance in the ambiguous region, with DQ increasing from 75.00\% to 100.00\%. Modification \#3 removes the penalty caused by the opposite assignment of the overlapping region, with SQ increasing from 84.45\% to 90.16\%. Modification \#4 compensates for the one-pixel boundary difference of the small instance, with SQ increasing from 84.45\% to 88.74\%. When all applicable modifications are combined, PQ, AJI, and SQ still do not reach 100\%. This is because the prediction assigns the overlapping region to a single instance, which is a genuine difference from the overlap-preserving ground truth rather than an evaluation artifact, and is therefore still penalized.

\subsection{Computational cost}
\label{sec:runtime}
Finally, to assess the computational overhead of the proposed modifications, we measured the average runtime per image under each configuration. For each image, the runtime was obtained as the mean over five repeated runs, and the reported values are averaged across all 665 images of the NuInsSeg dataset. The results are reported in Table~\ref{tab:runtime}. Modification \#2 (score normalization) is not included, as it is computed once for the entire dataset rather than per image, and its computational cost is therefore negligible.

\begin{table}[h]
	\centering
	\caption{Average runtime per image of the evaluation pipeline under the baseline and cumulatively modified configurations on the NuInsSeg dataset. Modification \#2 is not included, as it is computed once for the entire dataset.}
	\begin{tabular}{l c}
		\hline
		\textbf{Modifications} & \textbf{Runtime per image (s)} \\
		\hline
		Baseline   & 0.179 \\
		\#1        & 0.213 \\
		\#1,3      & 0.224 \\
		\#1,3,4    & 0.239 \\
		\hline
	\end{tabular}
	\label{tab:runtime}
\end{table}

As Table~\ref{tab:runtime} shows, the modifications add only modest computational overhead, with the fully modified pipeline requiring, on average, 0.06 more seconds per image than the baseline. This overhead is small compared to the inference time of typical DL-based nuclear instance segmentation models, and therefore NucEval can be used in practice without a substantial increase in evaluation time.

\subsection{Limitations and future studies}
\label{sec:limitaion}
Although we have proposed four modifications to enhance the evaluation pipeline for nuclear instance segmentation, a number of challenges necessarily limited the scope of our current investigation, leaving the door open for future research.

Our modification for addressing vague areas requires manual annotation of these regions, which have been mostly lacking in publicly available datasets and would require additional annotation effort. Outside of the studies discussed in Section~\ref{sec:vague areas}, the remaining publicly available datasets do not contain annotations for vague areas. Moreover, the vague area annotations in the NuInsSeg dataset are subject to the subjectivity of a single annotator. Therefore, repeat annotations by multiple expert annotators would be needed to further assess the reliability of the annotated vague areas.

Regarding score normalization, we note that nuclear count weighting is not the only valid aggregation strategy. Sparsely populated regions can be clinically important, and if patch- or region-level importance metadata were available, an importance-weighted aggregation would be more appropriate. However, to the best of our knowledge, such metadata are not available in current public nuclear instance segmentation datasets. Therefore, NucEval supports both normalized and non-normalized evaluation, allowing users to choose the aggregation strategy that best fits their application.

With respect to overlapping areas, while some datasets such as NuInsSeg and CryoNuSeg provide overlapping region annotations in their original raw annotation files, other datasets do not. Consequently, the proposed modification for handling overlapping areas cannot be applied to those datasets. In future studies, especially those involving the release of publicly available datasets for nuclear instance segmentation, the preservation of overlapping regions during the annotation process should be considered so that this modification can be broadly applied.

Regarding the border uncertainty zone, the current implementation applies a fixed zone width uniformly to all nuclei regardless of their size. However, in practice, boundary uncertainty may not be uniform across all instances. Larger nuclei with smoother boundaries may have less annotation uncertainty than smaller nuclei with irregular shapes. Future work could explore adaptive zone widths that scale with nuclear size or shape complexity (for example, by defining the zone width as a fraction of each instance's radius or perimeter). Nevertheless, providing the zone width as a hyperparameter in the current implementation gives users full control over implementation of this modification to the segmentation masks based on their specific application and dataset characteristics.

In this study, we used three task-specific models to show that the effect of the proposed modifications is model-agnostic. Evaluating additional model families, such as recent foundation models, would broaden the scope of the analysis. It should be noted, however, that task-specific models can still outperform foundation model-based approaches on specialized tasks. For example, in the PUMA challenge, task-specific approaches such as KongNet and a multi-stage auto-context approach outperformed methods based on foundation model backbones~\citep{schuiveling2025novel, lv2025kongnet, torbati2025multi}. Evaluating a wider range of models, including foundation models, with NucEval remains a direction for future work.

Finally, it should be noted that, while our experimental results are focused on nuclear instance segmentation, the proposed modifications are not inherently domain-specific. The NucEval pipeline can be applied to other instance segmentation tasks where similar evaluation challenges exist, such as nuclear instance segmentation and classification on a per-class basis, cell instance segmentation, or gland instance segmentation. More broadly, the concepts of ambiguous region exclusion, score normalization, and boundary uncertainty handling are relevant to any instance segmentation task, including applications beyond medical imaging, such as remote sensing.

\section{Conclusion}
In computational pathology, rigorous and reproducible evaluation of model performance is critical for comparing and selecting models for downstream clinical applications, including those based on nuclear segmentation. However, the performance evaluation of nuclear segmentation models  has received relatively little research attention relative to the importance of such models in computational pathology, resulting in significant limitations in current evaluation pipelines. In this study, we have identified four commonly encountered issues associated with nuclear instance segmentation evaluation and propose the following corresponding solutions: Modification \#1 to handle vague regions, Modification \#2 to normalize scores based on nuclear count, Modification \#3 to handle overlapping regions, and Modification \#4 to handle border uncertainty. Our experimental results across three datasets and three state-of-the-art models show that these modifications improve evaluation metrics in most cases, with Modification \#4 (border uncertainty zone) being the most universally impactful across all datasets and models. Modifications \#1 and \#2 show gains but are dataset-dependent, and Modification \#3 has a comparatively smaller impact on the results. Moreover, the modifications increase both inter- and intra-observer agreement between expert annotations on CryoNuSeg, and they can change the ranking of closely competing methods, as shown for top-ranked teams of the MoNuSAC challenge, indicating that the choice of evaluation pipeline can affect model selection. Importantly, all modifications are model-agnostic and can be selectively applied, depending on the availability of relevant annotations. By making our implementation publicly available and providing an easy-to-use, versatile NucEval function, we aim to make the evaluation pipeline for nuclear instance segmentation more reliable and transparent. This can lead to fairer model comparison and selection, and ultimately benefit the multitude of downstream clinical applications that directly or indirectly rely on accurate and precise nuclear instance segmentation.

\section*{CRediT authorship contribution statement}
\textbf{Amirreza Mahbod:} Conceptualization, Methodology, Software, Formal analysis, Investigation, Visualization, Writing - original draft, Writing - review \& editing, Funding acquisition.
\textbf{Ramona Woitek:} Supervision, Formal analysis, Writing - review \& editing.
\textbf{Jeanne Shen:} Supervision, Conceptualization, Formal analysis, Writing - review \& editing, Funding acquisition.

\section*{Funding}
This work was supported by Fulbright Austria (Fulbright Grant for Teaching, Research, Career Development, and Institutional Collaboration), as well as by the United States National Cancer Institute (NCI), National Institutes of Health (NIH) (R01 CA270437).

\section*{Declaration of Competing Interest}
The authors declare that they have no known competing financial interests that could have appeared to influence the work reported in this paper.

\section*{Data Availability}
The datasets used in this study are publicly available from previously published papers. The code developed for this study is available on GitHub, with the corresponding link provided in the manuscript.

\section*{Ethics Statement}
Ethical approval was not required for this study, as all datasets used are publicly available and were previously published.

\section*{Declaration of generative AI and AI-assisted technologies in the manuscript preparation process}
During the preparation of this work, the authors used ChatGPT (GPT 6.1 Sol) and Claude (Opus 5.5) to improve the language and readability of the manuscript. After using these tools, the authors reviewed and edited the content as needed and take full responsibility for the content of the published article.

\bibliographystyle{elsarticle-harv} 
\bibliography{refs.bib}

\appendix
\setcounter{figure}{0}
\setcounter{table}{0}
\section{Supplementary Information}

\subsection{Supplementary figures}

\begin{figure}[H]
	\centering
	\begin{tabular}{c}
		
		\includegraphics[width=\textwidth]{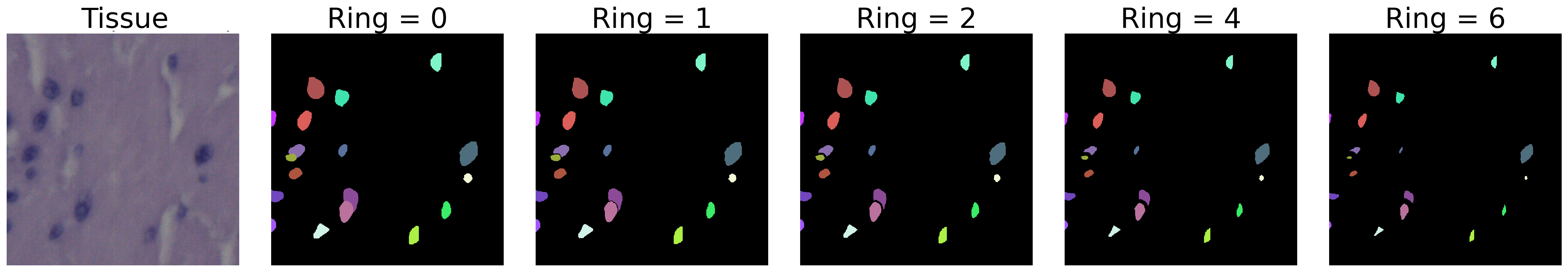} \\		
		\includegraphics[width=\textwidth]
		{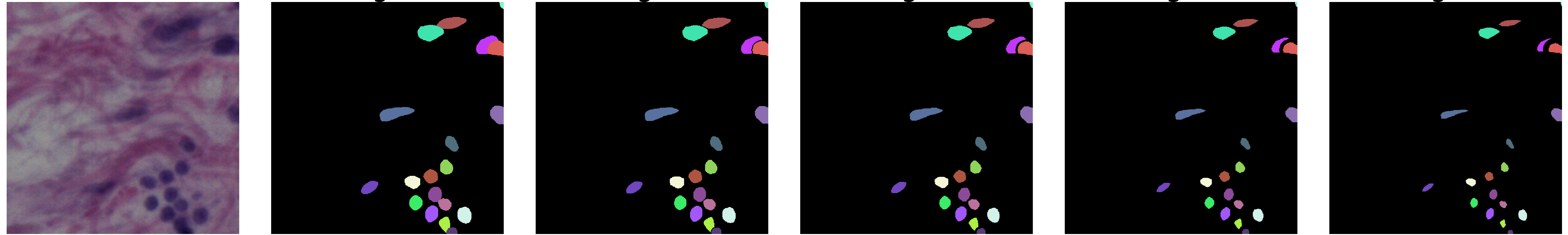} \\
		\includegraphics[width=\textwidth]
		{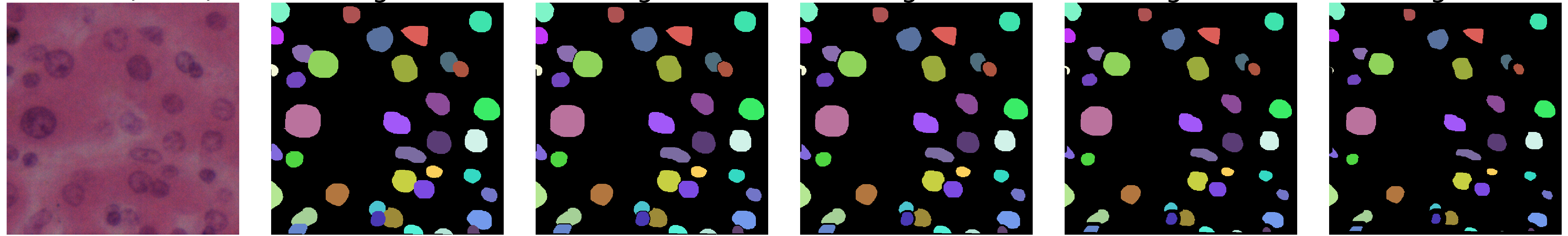} \\
		
	\end{tabular}
	\caption{Effect of zone size (ring) on the labeled segmentation masks. The example tissue images and corresponding annotations (ring = 0) are extracted from the NuInsSeg dataset~\citep{mahbod2024nuinsseg}. }
	\label{fig:example_ring}
\end{figure}

\begin{figure}[H]
	\centering
	\begin{subfigure}{0.48\textwidth}
		\centering
		\includegraphics[width=\linewidth]{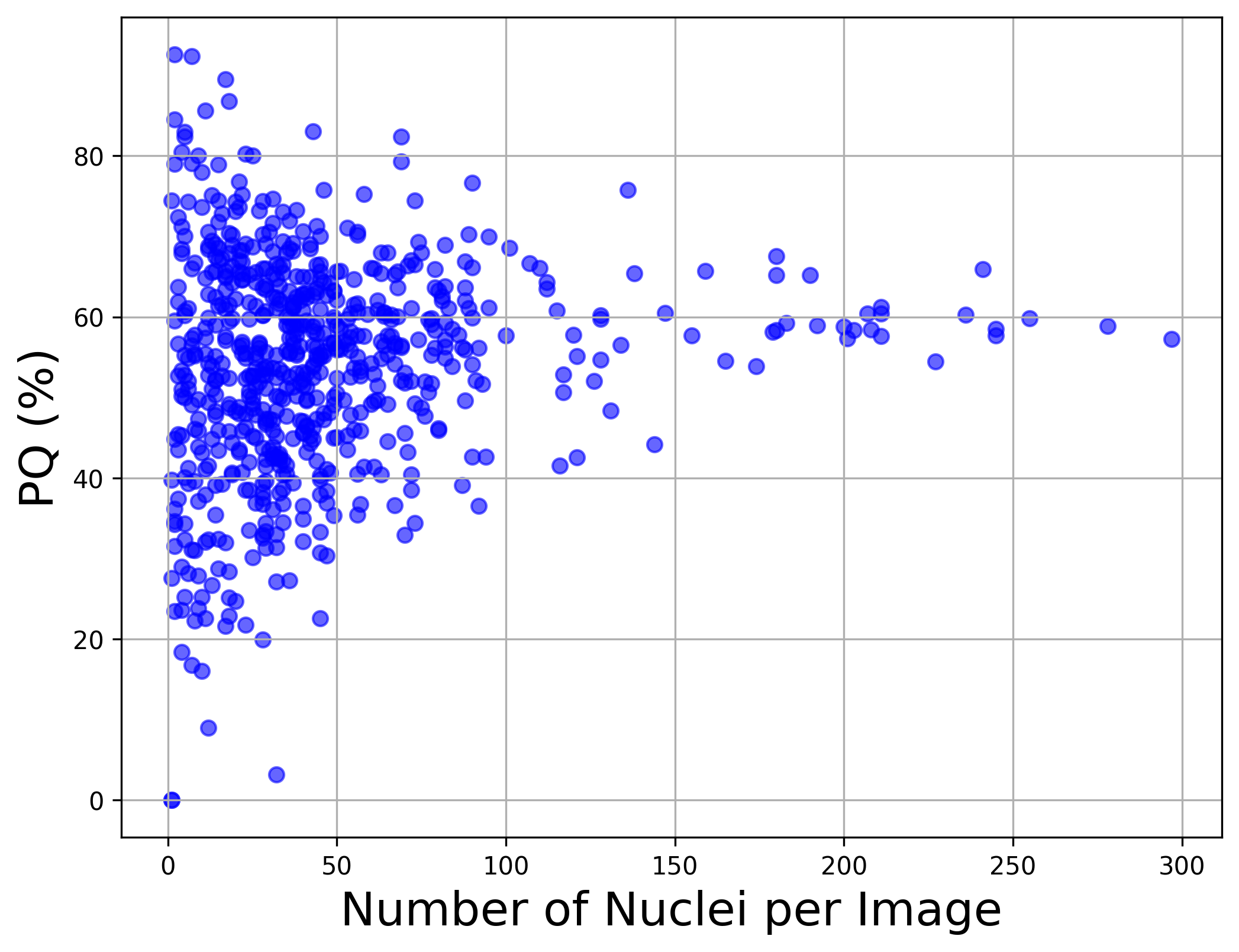}
		
	\end{subfigure}
	\hfill
	\begin{subfigure}{0.48\textwidth}
		\centering
		\includegraphics[width=\linewidth]{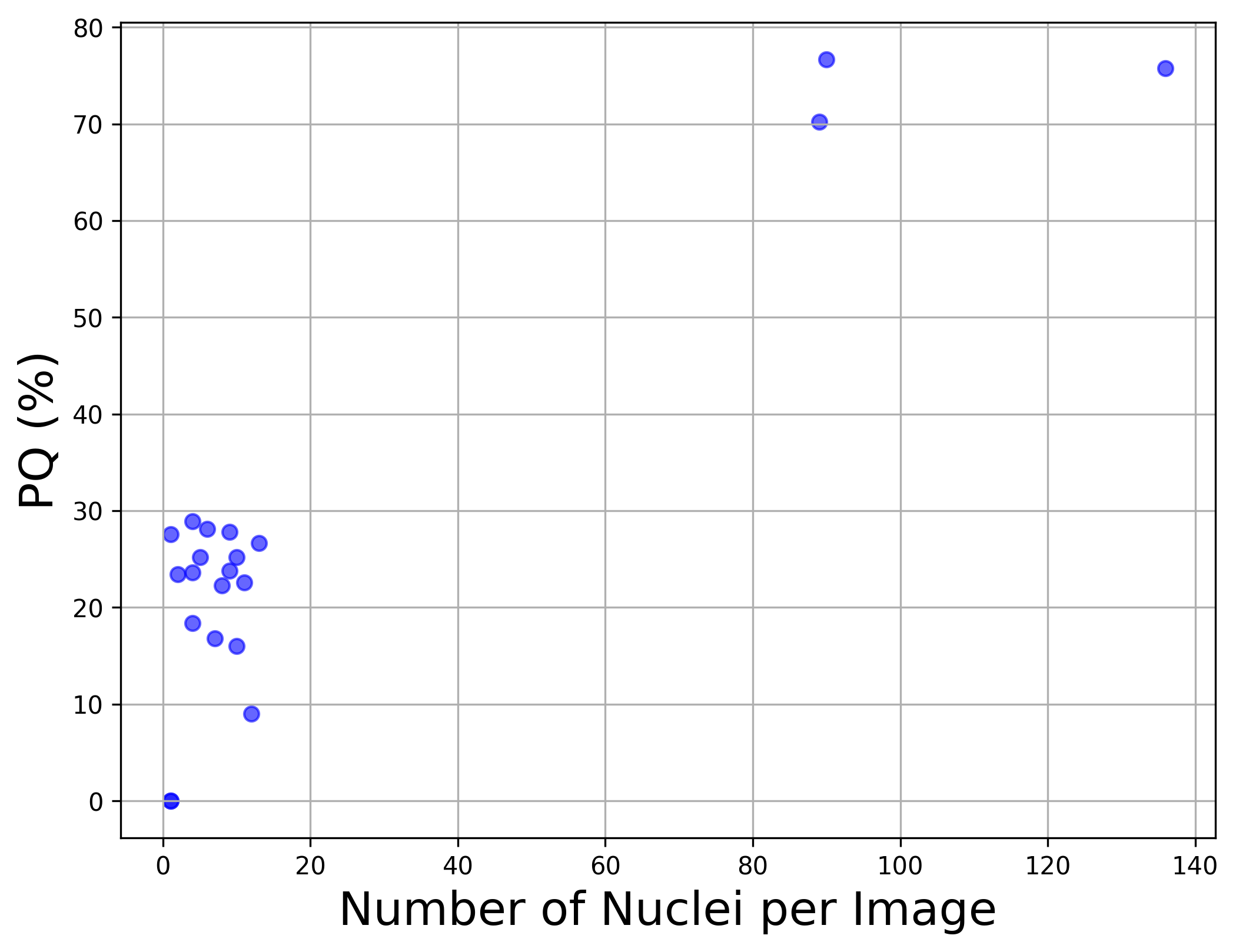}
		
	\end{subfigure}
	
	\caption{Scatterplot of PQ (\%) versus the number of nuclei per image for the full NuInsSeg dataset (left) and a selected subset (right), using the baseline HoVer-NeXt model. As observed in the left panel, the score variance decreases with increasing nuclear density, while the mean performance plateaus near 60\% for images containing more than approximately 100 nuclei.}
	\label{fig:scaterplot}
\end{figure}

\begin{figure}[H]
	\centering
	\includegraphics[width=0.6\linewidth]{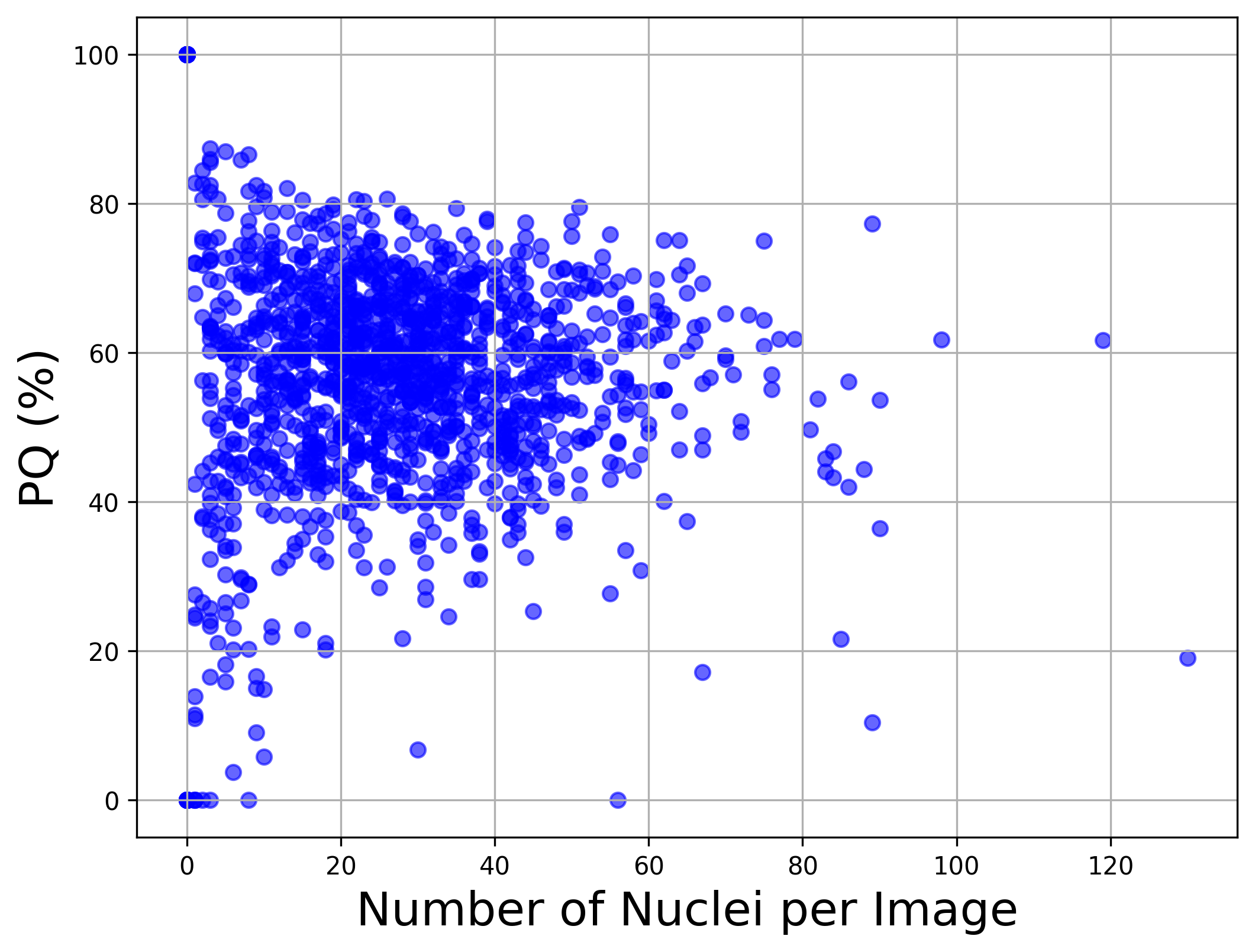}
	\caption{Scatterplot of PQ (\%) versus the number of nuclei per image for the PCNS dataset, using the baseline HoVer-NeXt model. }
	\label{fig:pq_aji_vs_nuclei_PCNS}
\end{figure}

In this section, we provide the supplementary figures referenced in the manuscript, including those illustrating the effect of zone size (Figure~\ref{fig:example_ring}), as well as the distribution of PQ scores for the NuInsSeg and PCNS datasets based on the number of nuclei per image (Figure~\ref{fig:scaterplot} and 
Figure~\ref{fig:pq_aji_vs_nuclei_PCNS}, respectively).


\subsection{Sensitivity to hyperparameters}
\label{sec:hyperparameter}

As mentioned in Section~\ref{sec:implementation_details} and Section~\ref{sec:vague areas}, we used a fixed zone size and threshold to remove ambiguous regions when reporting the results. We provide detailed results for various values of these two hyperparameters and their effect on the metrics in Table~\ref{tab:amb_threshold} and Table~\ref{tab:ring_size}. As shown in Table~\ref{tab:amb_threshold}, the best overall performance is achieved with $\texttt{overlap\_thresh\_amb} = 0.25$, although the differences across threshold values are relatively small, suggesting that the evaluation is not overly sensitive to this parameter. The extreme case of $\texttt{overlap\_thresh\_amb}$ = 0.01 (removing nearly all instances touching ambiguous regions) slightly degrades performance, indicating that overly aggressive removal discards useful instances. From Table~\ref{tab:ring_size}, it is evident that increasing the zone width improves the metrics (with the only exception being a slight decrease in the Dice score when the zone width increases from 4 to 6), since more boundary pixels are excluded from evaluation. However, this improvement should be interpreted with caution, as discussed in Section~\ref{sec:border_uncertain}, since excessively large zone widths reduce the effective evaluation area and may inflate the scores beyond what is meaningful. We selected a zone width of 1 as a conservative choice that accounts for estimated annotation uncertainty without discarding excessive evaluation area.

Throughout this study, we used an IoU matching threshold of 0.5, which is the standard choice in the nuclear instance segmentation literature~\citep{graham2019hover}. In addition, Table~\ref{tab:iou_sweep} reports PQ and DQ for IoU matching thresholds from 0.5 to 0.9, before and after applying each modification individually. All scores decrease as the matching threshold becomes stricter. Modification \#4 yields the largest PQ improvement at every threshold, and Modification \#1 improves both PQ and DQ at every threshold, whereas the small gains of Modifications \#2 and \#3 turn into slight decreases at the strictest thresholds (0.9 for Modification \#2, and 0.8 and 0.9 for Modification \#3).

\begin{table}[h]
	\centering
	\caption{Effect of ambiguity-aware instance removal on evaluation metrics. 
		An instance is removed if the ratio of its overlap with ambiguous regions exceeds a threshold $\tau$ ($\texttt{overlap\_thresh\_amb}$). 
		The baseline corresponds to no removal. 
		PQ: Panoptic Quality; AJI: Aggregated Jaccard Index; DQ: Detection Quality; SQ: Segmentation Quality. 
		All values are reported in \%, and results are obtained using the HoVer-NeXt model. }
	\label{tab:amb_threshold}
	\begin{tabular}{lccccccc}
		\toprule
		\textbf{Metric} & $\boldsymbol{\tau{=}0.99}$ & $\boldsymbol{\tau{=}0.75}$ & $\boldsymbol{\tau{=}0.5}$ & $\boldsymbol{\tau{=}0.25}$   & $\boldsymbol{\tau{=}0.01}$ & \textbf{Baseline} 
		\\
		\midrule
		\textbf{PQ}  & 55.45          & 56.69          & 56.77          & \textbf{56.80}   & 56.34  & 54.17 
		\\
		\textbf{AJI} & 60.75          & 60.96          & 61.03          & \textbf{61.06}   & 60.32  & 58.45 
		\\
		\textbf{Dice}& \textbf{83.82} & \textbf{83.82} & \textbf{83.82} & 83.81            & 83.06  & 82.59 
		\\
		\textbf{DQ}  & 73.11          & \textbf{74.98} & 74.79          & 74.83            & 74.22  & 72.14 
		\\
		\textbf{SQ}  & 75.13          & \textbf{75.17} & 75.16          & \textbf{75.17}   & 75.05  & 74.40 
		\\
		\bottomrule
	\end{tabular}
	
\end{table}

\begin{table}[h]
	\centering
	\caption{Effect of zone size on the performance. 
		PQ: Panoptic Quality; AJI: Aggregated Jaccard Index; DQ: Detection Quality; SQ: Segmentation Quality (size=0 represents the baseline). 
		All values are reported in \%, and results are obtained using the HoVer-NeXt model. }
	\label{tab:ring_size}
	
	\begin{tabular}{lccccccc}
		\toprule
		\textbf{Metric} & \textbf{size=0} & \textbf{size=1} & \textbf{size=2} & \textbf{size=4}  & \textbf{size=6}  \\
		\midrule
		\textbf{PQ}    &54.17  &60.01  &65.15  &70.60    &74.15    \\
		\textbf{AJI}   &58.45  &62.37  &65.77  &68.68    &69.11    \\
		\textbf{Dice}  &82.59  &85.46  &87.66  &88.86    &87.84    \\
		\textbf{DQ}    &72.14  &74.58  &76.38  &78.08    &79.68    \\
		\textbf{SQ}    &74.40  &79.78  &84.61  &89.81    &92.50   \\
		\bottomrule
	\end{tabular}
\end{table}
\begin{table}[h]
	\centering
	\caption{Effect of the IoU matching threshold on PQ and DQ before (baseline) and after applying each modification individually. 
		PQ: Panoptic Quality; DQ: Detection Quality. 
		All values are reported in \%, and results are obtained using the HoVer-NeXt model on the NuInsSeg dataset.}
	\label{tab:iou_sweep}
	
	\begin{tabular}{llccccc}
		\toprule
		\textbf{Metric} & \textbf{Configuration} & \textbf{0.5} & \textbf{0.6} & \textbf{0.7} & \textbf{0.8} & \textbf{0.9} \\
		\midrule
		\multirow{5}{*}{\textbf{PQ}} & Baseline         & 54.17 & 48.53 & 39.76 & 25.56 & 4.81 \\
		                             & Modification \#1 & 56.80 & 51.37 & 42.67 & 28.67 & 6.05 \\
		                             & Modification \#2 & 55.96 & 50.53 & 42.11 & 27.06 & 4.51 \\
		                             & Modification \#3 & 54.70 & 49.03 & 40.20 & 24.65 & 4.29 \\
		                             & Modification \#4 & 60.01 & 55.65 & 49.46 & 31.86 & 8.46 \\
		\midrule
		\multirow{5}{*}{\textbf{DQ}} & Baseline         & 72.14 & 61.95 & 48.54 & 29.67 & 5.24 \\
		                             & Modification \#1 & 74.83 & 65.00 & 51.71 & 33.14 & 6.54 \\
		                             & Modification \#2 & 74.18 & 64.34 & 51.44 & 31.47 & 4.91 \\
		                             & Modification \#3 & 73.18 & 62.87 & 49.34 & 28.70 & 4.67 \\
		                             & Modification \#4 & 74.58 & 66.67 & 57.17 & 34.65 & 8.74 \\
		\bottomrule
	\end{tabular}
\end{table}
\FloatBarrier
\subsection{Complete cumulative results}
In this section, we provide the complete cumulative results for all metrics and for each dataset, namely the full NuInsSeg dataset (Table~\ref{tab:nuinsseg_entire_cumulative}), the NuInsSeg subset (Table~\ref{tab:nuinsseg_subset_cumulative}), CryoNuSeg (Table~\ref{tab:cryonuseg_entire_cumulative}), and PCNS (Table~\ref{tab:pcns_entire_cumulative}), respectively.

\begin{table}[h]
	\centering
	\caption{Effect of cumulative modifications on the entire NuInsSeg dataset.
		PQ: Panoptic Quality; AJI: Aggregated Jaccard Index;
		DQ: Detection Quality; SQ: Segmentation Quality.
		All values are reported in \%.}
	\begin{tabular}{ll ccccc}
		\hline
		\textbf{Model} & \textbf{Modifications} & \textbf{PQ} & \textbf{AJI} & \textbf{DICE} & \textbf{DQ} & \textbf{SQ} \\
		\hline
		\multirow{5}{*}{\textbf{HoVer-Net}}
		& Baseline           & 53.40 & 57.50 & 80.29 & 70.38 & 75.17 \\
		&  \#1            & 56.17 & 60.75 & 81.31 & 73.28 & 75.82 \\
		&  \#1,2       & 57.05 & 59.63 & 83.73 & 74.15 & 76.64 \\
		&  \#1,2,3  & 58.18 & 60.77 & 83.75 & 76.61 & 75.68 \\
		&  \#1,2,3,4 & 64.07 & 64.91 & 86.60 & 78.64 & 81.24 \\
		\hline
		\multirow{5}{*}{\textbf{HoVer-NeXt}}
		& Baseline           & 54.17 & 58.45 & 82.59 & 72.14 & 74.40 \\
		& \#1            & 56.80 & 61.06 & 83.81 & 74.83 & 75.17 \\
		&  \#1,2       & 58.29 & 61.52 & 85.73 & 76.51 & 75.88 \\
		&  \#1,2,3  & 58.45 & 61.75 & 85.66 & 77.35 & 75.31 \\
		&  \#1,2,3,4 & 64.92 & 66.41 & 88.76 & 79.90 & 81.00 \\
		\hline
		\multirow{5}{*}{\textbf{CellViT}}
		& Baseline           & 56.44 & 60.03 & 81.92 & 74.50 & 74.83 \\
		&  \#1            & 59.72 & 63.31 & 83.17 & 77.95 & 75.58 \\
		&  \#1,2       & 60.92 & 63.08 & 85.32 & 79.33 & 76.51 \\
		&  \#1,2,3  & 61.93 & 64.02 & 85.25 & 81.12 & 76.08 \\
		&  \#1,2,3,4 & 68.44 & 68.87 & 88.40 & 83.27 & 81.96 \\
		\hline
	\end{tabular}
	\label{tab:nuinsseg_entire_cumulative}
\end{table}

\begin{table}[h]
	\centering
	\caption{Effect of cumulatively applied evaluation protocol modifications
		on the subset of the NuInsSeg dataset.
		PQ: Panoptic Quality; AJI: Aggregated Jaccard Index;
		DQ: Detection Quality; SQ: Segmentation Quality.
		All values are reported in \%.}
	\begin{tabular}{ll ccccc}
		\hline
		\textbf{Model} & \textbf{Modifications} & \textbf{PQ} & \textbf{AJI} & \textbf{DICE} & \textbf{DQ} & \textbf{SQ} \\
		\hline
		\multirow{5}{*}{\textbf{HoVer-Net}}
		& Baseline                & 28.68 & 34.42 & 54.65 & 38.79 & 62.79 \\
		&  \#1                 & 40.03 & 46.92 & 63.74 & 49.84 & 65.65 \\
		&  \#1,2            & 60.21 & 63.90 & 82.20 & 73.38 & 78.96 \\
		&  \#1,2,3       & 60.90 & 64.34 & 82.22 & 75.62 & 77.37 \\
		&  \#1,2,3,4  & 67.13 & 69.15 & 85.10 & 77.26 & 83.52 \\
		\hline
		\multirow{5}{*}{\textbf{HoVer-NeXt}}
		& Baseline                & 28.00 & 36.77 & 59.01 & 38.06 & 64.87 \\
		&  \#1                 & 40.35 & 49.15 & 69.98 & 51.14 & 67.36 \\
		&  \#1,2            & 64.95 & 68.14 & 85.19 & 79.27 & 79.02 \\
		&  \#1,2,3       & 64.18 & 67.51 & 85.13 & 79.02 & 78.65 \\
		&  \#1,2,3,4  & 70.61 & 73.08 & 88.04 & 80.40 & 85.20 \\
		\hline
		\multirow{5}{*}{\textbf{CellViT}}
		& Baseline                & 29.07 & 36.93 & 56.30 & 38.57 & 63.44 \\
		&  \#1                 & 43.45 & 51.78 & 66.72 & 53.46 & 66.42 \\
		&  \#1,2            & 65.66 & 69.43 & 84.61 & 79.87 & 79.26 \\
		&  \#1,2,3       & 66.15 & 69.51 & 84.65 & 81.48 & 78.42 \\
		&  \#1,2,3,4  & 72.26 & 75.29 & 87.81 & 81.89 & 85.35 \\
		\hline
	\end{tabular}
	\label{tab:nuinsseg_subset_cumulative}
	
\end{table}

\begin{table}[h]
	\centering
	\caption{Effect of cumulative modifications on the CryoNuSeg dataset. Modification \#1 is not applicable to CryoNuSeg, as vague area annotations are not available for this dataset.
		PQ: Panoptic Quality; AJI: Aggregated Jaccard Index;
		DQ: Detection Quality; SQ: Segmentation Quality.
		All values are reported in \%.}
	\begin{tabular}{ll ccccc}
		\hline
		\textbf{Model} & \textbf{Modifications} & \textbf{PQ} & \textbf{AJI} & \textbf{DICE} & \textbf{DQ} & \textbf{SQ} \\
		\hline
		\multirow{5}{*}{\textbf{HoVer-Net}}
		& Baseline    & 47.84    & 53.67      & 80.33   & 63.58    & 74.81 \\
		&  \#2      & 50.47    & 55.20      & 82.74   & 66.51    & 75.45 \\
		&  \#2,3    & 49.31    & 55.19      & 81.15   & 67.69    & 72.43  \\
		&  \#2,3,4  & 59.22    & 61.84      & 86.09   & 72.28    & 81.44 \\
		\hline
		\multirow{5}{*}{\textbf{HoVer-NeXt}}
		& Baseline    & 48.12    & 51.21    & 80.31   & 63.85    & 75.02 \\
		&  \#2      & 49.96    & 52.71    & 82.16   & 65.74    & 75.61  \\
		&  \#2,3    & 49.97    & 52.94    & 80.85   & 67.80    & 73.38 \\
		&  \#2,3,4  & 58.92    & 59.25    & 85.85   & 71.58    & 81.95 \\
		\hline
		\multirow{5}{*}{\textbf{CellViT}}
		& Baseline    & 45.79     & 51.65       &  79.35   &  62.32    & 73.11  \\
		
		&  \#2      & 47.35     & 52.63       & 81.15    & 64.18     & 73.40\\
		&  \#2,3    & 49.21     & 54.42       &  81.20   & 67.03     & 72.96  \\
		&  \#2,3,4  & 58.55     & 60.70       & 85.93    & 71.25     & 81.67  \\
		\hline
	\end{tabular}
	\label{tab:cryonuseg_entire_cumulative}
\end{table}


\begin{table}[h]
	\centering
	\caption{Effect of cumulative modifications on the PCNS dataset. Modifications \#1 and \#3 are not applicable to PCNS, as vague area annotations and ROI annotations for each instance are not available for this dataset.
		PQ: Panoptic Quality; AJI: Aggregated Jaccard Index;
		DQ: Detection Quality; SQ: Segmentation Quality.
		All values are reported in \%.}
	\begin{tabular}{ll ccccc}
		\hline
		\textbf{Model} & \textbf{Modifications} & \textbf{PQ} & \textbf{AJI} & \textbf{DICE} & \textbf{DQ} & \textbf{SQ} \\
		\hline
\multirow{5}{*}{\textbf{HoVer-Net}}
& Baseline    & 58.04 & 60.19 & 79.71 & 74.59 & 76.07 \\

&  \#2      & 58.27 & 59.91 & 81.82 & 75.37 & 76.65 \\

&  \#2,4  & 66.37 & 65.64 & 85.87 & 77.95 & 84.47 \\
\hline
\multirow{5}{*}{\textbf{HoVer-NeXt}}
& Baseline    & 56.74 & 57.76 & 79.87 & 73.14 & 76.09 \\

&  \#2      & 56.69 & 57.70 & 81.89 & 73.58 & 76.45 \\

&  \#2,4  & 64.40 & 62.99 & 85.88 & 76.12 & 84.01 \\
\hline
\multirow{5}{*}{\textbf{CellViT}}
& Baseline    & 60.75 & 62.42 & 81.20 & 77.50 & 76.88 \\

&  \#2      & 60.71 & 62.57 & 83.03 & 78.22 & 77.10 \\

&  \#2,4  & 69.10 & 68.64 & 87.22 & 80.73 & 85.04 \\
\hline
	\end{tabular}
	\label{tab:pcns_entire_cumulative}
\end{table}

\FloatBarrier

\end{document}